\documentclass[lettersize,journal]{IEEEtran}
\usepackage{amsmath,amsfonts}
\usepackage{array}
\usepackage[caption=false,font=normalsize,labelfont=sf,textfont=sf]{subfig}
\usepackage{textcomp}
\usepackage{stfloats}
\usepackage{url}
\usepackage{verbatim}
\usepackage{graphicx}
\usepackage{cite}
\usepackage{rotating}
\usepackage{amssymb}

\usepackage{amsmath}
\usepackage{algorithm}
\usepackage[noend]{algpseudocode}
\usepackage{textgreek}
\usepackage{xfrac}
\usepackage{colortbl}
\usepackage{threeparttable, tablefootnote}
\usepackage{booktabs}
\usepackage{multirow}
\usepackage{tabularx}
\usepackage{longtable}
\usepackage{tablefootnote}
\newcolumntype{K}[1]{>{\centering\arraybackslash}p{#1}}
\newcolumntype{L}[1]{>{\arraybackslash}p{#1}}
\usepackage{lscape} % <-- HERE
\usepackage{mathtools}

\usepackage{graphicx}

\usepackage{hyperref}

\usepackage{xcolor}
\definecolor{amber}{rgb}{1.0, 0.49, 0.0}
\definecolor{cadmiumgreen}{rgb}{0.0, 0.42, 0.24}

\usepackage{amsthm}
\usepackage{tikz}

\newtheoremstyle{styleth}%
{3pt}% Space above
{3pt}% Space below 
{}% Body font
{}% Indent amount
{\bfseries\color{amber}}% Theorem head font
{}% Punctuation after theorem head
{.5em}% Space after theorem head
{}% Theorem head spec (can be left empty, meaning ‘normal’)
\theoremstyle{styleth}

\newtheoremstyle{styledef}%
{3pt}% Space above
{3pt}% Space below 
{}% Body font
{}% Indent amount
{\bfseries\color{cadmiumgreen}}% Theorem head font
{}% Punctuation after theorem head
{0.5em}% Space after theorem head
{}% Theorem head spec (can be left empty, meaning ‘normal’)
\theoremstyle{styledef}

\usepackage[most]{tcolorbox}

\newtcolorbox{definition_}{
enhanced,
boxrule=0pt,frame hidden,
borderline west={2pt}{0pt}{blue!55!black},
colback=blue!5!white,
sharp corners
}

\begin{document}

\algnewcommand\algorithmicforeach{\textbf{for each}}
\algdef{S}[FOR]{ForEach}[1]{\algorithmicforeach\ #1\ \algorithmicdo}

\title{An experimental survey and Perspective View on Meta-Learning for Automated Algorithms Selection and Parametrization}

%\author{IEEE Publication Technology,~\IEEEmembership{Staff,~IEEE,}
        % <-this % stops a space

 %}

\author{
    \IEEEauthorblockN{Moncef Garouani\IEEEauthorrefmark{1}}\\
    \IEEEauthorblockA{\IEEEauthorrefmark{1}IRIT, UMR 5505 CNRS, Université Toulouse Capitole, UT1, Toulouse, France\\moncef.garouani@irit.fr}\\
}


\maketitle

\begin{abstract}
Considerable progress has been made in the recent literature studies to tackle the Algorithms Selection and Parametrization\,(ASP) problem, which is diversified in multiple meta-learning setups. Yet there is a lack of surveys and comparative evaluations that critically analyze, summarize and assess the performance of existing methods. In this paper, we provide an overview of the state of the art in this continuously evolving field. The survey sheds light on the motivational reasons for pursuing classifiers selection through meta-learning. In this regard, Automated Machine Learning\,(AutoML) is usually treated as an ASP problem under the umbrella of the democratization of machine learning. Accordingly, AutoML makes machine learning techniques accessible to domain scientists who are interested in applying advanced analytics but lack the required expertise. It can ease the task of manually selecting ML algorithms and tuning related hyperparameters. We comprehensively discuss the different phases of classifiers selection based on a generic framework that is formed as an outcome of reviewing prior works. Subsequently, we propose a benchmark knowledge base of \textit{4\,millions} previously learned models and present extensive comparative evaluations of the prominent methods for classifiers selection based on 08 classification algorithms and 400 benchmark datasets. The comparative study quantitatively assesses the performance of algorithms selection methods along while emphasizing the strengths and limitations of existing studies.
\end{abstract}


\section{Introduction}

Machine Learning\,(ML) solutions often require massive data analysis. Nowadays, The increasing data availability and advanced data analytics have fueled the ML-based solutions to contribute in humans relief from many risky, repetitive and tedious activities. ML became a vital part in many aspects of daily life\,\cite{Garouani_sncs}. To give just a few examples, ML can suggest active user that which books or newspapers should be read\,\cite{caldarelliSignalBasedApproachNews2016}, what movies to watch\,\cite{biancalanaContextawareMovieRecommendation2011}, what music to listen to\,\cite{onoriComparativeAnalysisPersonalityBased2016},~etc. 
ML approaches has been employed to develop systems that are able to recommend which places\,(e.g., cultural and artistic attractions\,\cite{sansonettiEnhancingCulturalRecommendations2019}) to visit\,\cite{sansonettiPointInterestRecommendation2019} and the best itinerary to reach there\,\cite{fogliExploitingSemanticsContextaware2019}. The ML solutions have been also proven to be very efficient in real-time systems such as self-driving cars\,\cite{kulkarniTrafficLightDetection2018} or predictive maintenance in Industry 4.0\,\cite{CoDIT_choaib,edas}, etc. However, along with these global applications, there is a widespread awareness that, given a specific problem, the process of designing and implementing a truly effective and efficient ML system requires considerable knowledge and effort by highly specialized data scientists and domain experts.

 %Each algorithm is intrinsically optimized and its performance on a particular task depends on how well its embedded fixed bias match the problem. Hence, there is no single algorithm that can learn all the tasks efficiently and every algorithm can perform better only on limited number of tasks. This phenomenon is also called performance complementarity\,\cite{khanLiteratureSurveyEmpirical2020}, and is also been confirmed by the well known \textit{No Free Lunch} theorem\,\cite{wolpertNoFreeLunch1997}.

The utility of ML algorithms have been utilized in multiple innovative ways. But yet, in most of cases these require manual interventions of human experts for intermediate tasks like data pre-processing, features engineering, the comparisons of several possible ML algorithms\,\cite{Garouani2022a}. Moreover, these algorithms are configured with hyperparameters\,(HPs) that can affect their predictive performance. Although, the configuration of HPs in optimal manner could help to discover the more accurate models but the optimization process to find the suitable HPs settings involves the resource intensive activities\,\cite{10849387}. Additionally, these optimization processes does not always yield the better performing results than the default HPs settings provided by ML packages and tools\,\cite{khan_2020}. It can be useful to know in advance the which algorithms is best suited for a given task in order to increase the performance or reduce the computational costs along with necessary tuning of HPs settings. Hence, there is an increasing need for automated support solutions that can be used without significant human interventions, in contrast to the increasing complexity of such processes.

Many studies have demonstrated the effectiveness of meta-learning when addressing the Algorithms Selection and Parameterization. In this context, the applications of AutoML with the use of Meta-Learning\,(MtL) have the potential to facilitate the efficiency of machine learning solutions\,\cite{garouaniAutomationIndustrialData2021}. This can help to alleviate the repetitive, time-consuming, and resource-intensive tasks that data scientists and practitioners may encounter. It can thus result in an increased democratization of machine learning algorithms. By automating these tasks, data scientists can allocate their time to more important responsibilities, such as problem formulation and analysis instead of fine tuning of HPs settings. The eventual objective of this research work is to provide support to analysts in selecting the most appropriate learning algorithm for a classification problem, thereby eliminating the need for systematic experimentation with various learning algorithms and HPs configurations.

For this purpose, a large number of methods are proposed for ML algorithms recommendation. Yet, there still lacks a comprehensive survey to be conducted in order to critically analyze, summarize, and identify the current and future challenges of meta-learning-based algorithms selection and parametrization methods. Furthermore, as can be expected, the extensive comparative studies must evaluate the performance of prominent algorithms selection methods, using the meta-learning, along with highlighting their strengths and weaknesses. The main reason for this literature gap is primarily due to the lack of reference knowledge bases that are the core of each MtL system and approach. To bridge the gap, our goal for this survey is to\,:
\begin{itemize}
    \item provide the reader with a better understanding of the algorithms selection and configuration problems,
    \item survey the meta-learning-based algorithms selection and parametrization methods,
    \item  guide the interested researchers towards building and improving MtL based AutoML systems according to currently available techniques and approaches, and
    \item finally propose a large and open-source knowledge base of more than \textit{4 millions} previous learning experiences. It could be used as a reference base for new methods and studies.
\end{itemize}

In this paper, the pertinent literature is examined, and the findings are presented that are subjected to critical discussions. In order to achieve this goal, as an initial step, this paper provides a comprehensive survey that presents a detailed analysis of significant dimensions of meta-learning for algorithm selection, and aims to address the research inquiries formulated on three fundamental dimensions: meta-\textit{features}, \textit{meta-models}, and \textit{meta-targets}. 

In this regard, the paper is organized as follows. Section\,\ref{sec:ASP} gives definitions and background information on the ASP. In Section\,\ref{sec:automl}, we review the definitions of AutoML and meta-learning as given in scientific literature, focusing on the categorization of the MtL techniques, based on the type of meta-data they leverage, from the most general to the most task-specific. Practical considerations arising when designing a meta-learning system are discussed in Sect.\,\ref{mtl_sec}, while open research directions are listed in Sect.\,\ref{Summary_sec}. The process of constructing the knowledge base is described in section\,\ref{sec:kb}, to report afterward in section\,\ref{sec:experiments} the benchmarking experimentations and results. Finally, open research directions are listed in Section\,\ref{sec:future_directions}.

\section{ML algorithms selection and parametrization}
\label{sec:ASP}

The recent literature studies reveals that the field of machine learning has made notable progress, particularly with the emergence of methodologies like deep learning  and the development of large-scale models. It has been evidenced by achievements in various domains such as robotics\,\cite{koberReinforcementLearningRobotics2013}, healthcare\,\cite{greenspanGuestEditorialDeep2016}, and autonomous driving\,\cite{chenDeepDrivingLearningAffordance2015}. However, the creation of such solutions demands making of use of substantial resources, both in terms of computational power and human expertise. The application of machine learning to real-world problems entails a multi-stage process that requires considerable human effort, ranging from data acquisition to model deployment\,\cite{10.1007/978-3-031-48232-8_42}. In this section, we address the issue of ML algorithm selection and parameterization, as well as the various methods employed to solve this problem.

\subsection{The algorithms selection problem}

It is observed that the field of machine learning has undergone consistent evolution. These advancements provided a range of models and algorithms to address supervised, semi-supervised, and unsupervised tasks. The design and development of a classification problem is a time and resource intensive endeavor, that involves several phases that necessitate technical knowledge usually held by expert analysts.

It starts with the identification of a particular learning problem and that continues with the practitioners choice of suitable learning tools to solve it. These tools may target various aspects of the machine learning pipeline, such as \textit{data preparation},\textit{ feature engineering}, \textit{model selection}, and \textit{hyperparameter tuning or optimization}. In this regard the quality of data plays a crucial role to achieve high-performance solutions. However, it need to delve into the coherent data collection, but the primary responsibility of data analyst is to select the most appropriate learning method\,(whether it be supervised or unsupervised learning). This task must be conducted in accordance with certain performance measures and within the constraints imposed by the application. The data analyst must identify the method that best corresponds to the morphology and specific characteristics of the given problem. This selection constitutes one of the most challenging problems, as there is no model or algorithm that outperforms all others, regardless of the specific problem characteristics, as has been observed in various empirical comparisons\,\cite{Garouani_WISE2024}.

\begin{figure}[h!]
	\centering
	\resizebox{1\hsize}{!}{\includegraphics{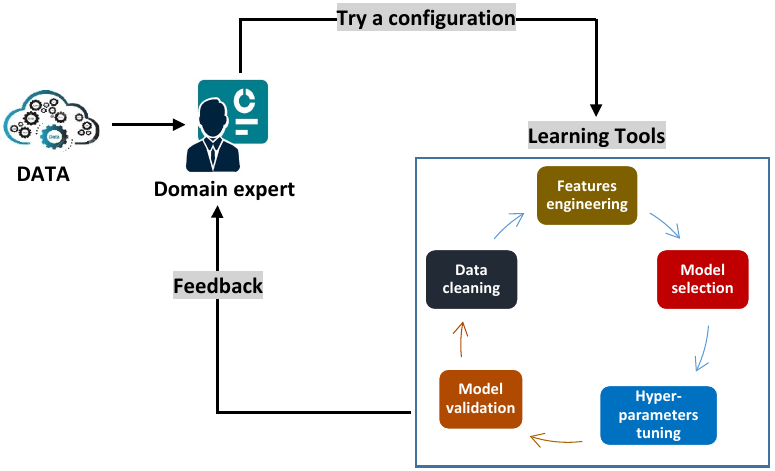}}
	\caption{The process of ML models selection for knowledge discovery.}
	\label{fig:humain_process}       % Give a unique label
\end{figure}

An overview of the process cycle as it is conventionally adopted by the data analysts is presented in Figure\,\ref{fig:humain_process}. The data analyst has a set of learning tools available, from which initially some can be chosen for evaluating the specific problem. This selection process may be based on prior knowledge of the problem, which involves the algorithms selections that have characteristics to best match those of the dataset, or the analyst's personal preferences for some specific learning algorithms. Evaluation of the algorithms typically involves extensive experimentation, which requires repeated executions of the selected algorithm(s).

In order to achieve high learning performance, the analyst initially attempts to configure the learning tools based on personal experience or intuition about the data and underlying tools. For this sake, based on the feedback received about the performance of the learning tools, the practitioner may adjust the configuration with the aim of improving the performance. This process is often iterative and involves trial-and-error approach until the desired performance is achieved or until the computational budget is exhausted.

\subsection{Hyperparameters tuning}
The task of selecting an appropriate algorithm or family of algorithms is crucial to ensure superior performance on a specific combination of datasets and evaluation metrics\,\cite{pria,cohen-shapiraAutoGRDModelRecommendation2019}. In machine learning, algorithms typically consist of two types of parameters, namely, the \emph{ordinary parameters} that are learned and optimized by the model automatically during the learning phase, and the \emph{hyperparameters},(discrete or continuous) that must be set manually prior to model training,(as illustrated in Table~\ref{tab:conf1}).

\begin{table}[h!]
	\caption{Configuration space of some classification algorithms.}
	\label{tab:conf1}  
	\begin{tabular}{lcc}
		\hline
		\noalign{\smallskip}
		ML Algorithm           & Nb. of ordinary parameters & Nb. of hyperparameters \\
		\noalign{\smallskip}\hline\noalign{\smallskip}
		SVM  & 2                      & 5                  \\
		Decision Tree          & 1                      & 3                  \\
		Random Forest          & 2                      & 4                  \\
		Logistic Regression    & 4                      & 6                  \\
		\noalign{\smallskip}\hline                                              
	\end{tabular}
\end{table}

\begin{figure}[h!]
	\centering
	\resizebox{1\hsize}{!}{\includegraphics{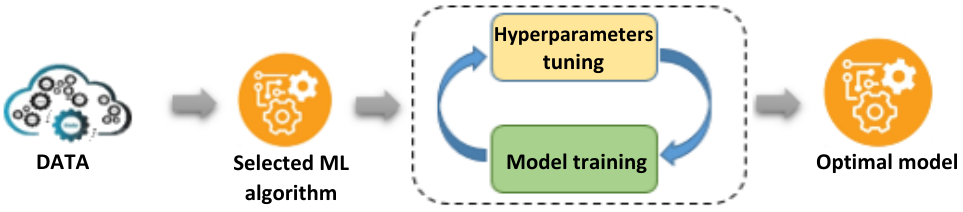}}
	\caption{General view of the hyperparameters tuning schema.}
	\label{fig:HP_tuning_process}       % Give a unique label
\end{figure}

Selecting feasible ML algorithms and tuning their hyperparameters is a major research challenge when approaching a new problem. Due to the nature of ML algorithms often being used as "black boxes," their performance is impacted by various characteristics of both the datasets and the algorithm's hyperparameters\,\cite{feurerEfficientRobustAutomated2015a}. As a result, the process of selecting and configuring the most appropriate algorithm(s) is both error-prone and time-consuming due to the numerous configurations that must be established. Figure\,\ref{fig:HP_tuning_process} provides an illustration of the general schema for hyperparameter tuning.

The process of Hyperparameters\,(HP) tuning is often done manually by experts who progressively adjust a grid of values within a desired range. However, theoretically, selecting the optimal HP values involves an exhaustive search over all possible combinations of HP subsets, which can become impractical as the number and types of HPs increases. The machine learning community often settles for searching a reduced space of HPs, rather than the complete space,\cite{bergstraRandomSearchHyperparameter2012} to address this issue in a better manner.
%However, literature provides a set of techniques ranging from simple to complex, able to accomplish this task.

\subsection{Definitions}
The hyperparameters tuning process is usually a black-box optimization problem. Its objective function is associated with the predictive performance of the model that is induced by some ML algorithm.

The described process can be defined in formal terms as follows\,\cite{garouthesis2022}:

\begin{definition_} 
	\textbf{DEFINITION 1. } Let $\mathcal{H}=\{H_{1}, \dots, H_{n}\}$ be the HPs space of an algorithm $A^{(i)}~\in~\mathcal{A}$, where $\mathcal{A}$ is the set of learning algorithms space. Each $A^{(i)*}~\in~\mathcal{A}$  represents a tuned version of $A^{(i)}$ and can be usually defined by a set of constraints.
\end{definition_}

\begin{definition_} 
	\textbf{DEFINITION 2. } Let $\mathcal{D}$ be a dataset divided into disjoint training denoted as $D_{train}$, and validation sets denoted as $D_{validation}$. The function  $\mathcal{L}$  measures the predictive performance of the model, as below:
	
	$\mathcal{L}:~A^{(i)}~\times H_{n} \times D \rightarrow \mathbb{R}$  
	
	Where, the predictive performance is induced by the algorithm $A^{(i)}$ with an hyperparameters configuration $H_n\in \mathcal{H} $  on the dataset $ D $. 
	
	Consequently, it can be observed that the higher values of  $\mathcal{L}(A^{(i)}, H_{n}, D_{train}, D_{validation})$ mean higher predictive performance, without loss of generality.
\end{definition_}

\begin{definition_}
	\textbf{DEFINITION 3. } Given $A^{(i)} \in \mathcal{A}$, $\mathcal{H}$ and $\mathcal{D}$,\,(from the definition 1 and definition 2), the goal of the algorithm selection and parametrization problem is to find the $A^{(i)*}$ that minimizes or maximizes the $\mathcal{L}$ on $\mathcal{D}$ such that\,:	\begin{center} $	A^{(i)*} \in \underset{A^{(i)} \in \mathcal{A}, H_{n} \in \mathcal{H}}{argmin\,\,} \mathcal{L}(A^{(i)}, H_{n}, D_{train}, D_{validation})$\end{center}
\end{definition_}

The optimization process of Combined Algorithms Selection and Hyperparameters\,(CASH) involves the identification of the most suitable algorithm\,($A^{(i)}~\in~\mathcal{A}$) along with its related hyperparameters configuration\,($H_n$) from a range of feasible configurations\,($\mathcal{H}$). This range of feasible configurations defines the search space, and it consists of either \emph{discrete} or \emph{continuous}  hyperparameters in a structured or unstructured manner\,\cite{bergstraRandomSearchHyperparameter2012,feurerInitializingBayesianHyperparameter2015}. Discrete hyperparameters include choices of activation function, number of layers in a neural network, or the number of trees in a random forest, etc. Continuous hyperparameters include among others learning rate or momentum for stochastic gradient descent, degree of regularization to be applied to the training objective, and scale of a kernel similarity function for kernel classification. Structured search spaces may include those with conditional hyperparameters\,(e.g., where the relevant hyperparameters depend on the choice of supervised learning method) and other tree type search spaces\,\cite{olsonEvaluationTreebasedPipeline2016}.

Various performance measures\,(e.g. Accuracy, AUC, Recall) can influence the optimization of the HPs values. These can also be defined by multi-objective criteria. However, there are several aspects that can make the tuning more difficult, including the fact that
HPs configurations that perform better on a particular dataset may not lead to high predictive performance for other datasets. In addition, HPs values are often depend on one another\,(as is the case of SVM and LR). Hence, independent tuning of these HPs may not result in an optimal set of HPs values. Furthermore, the exhaustive evaluation of HPs configuration can be extremely time and resource consuming and it can result in expensive computational costs.

There are several methods for the selection of configurations to evaluate their optimal performance within a given search space. The two most commonly used methods are \textit{Grid Search}\,(GS) and \textit{Random Search}\,(RS)\,\cite{andradottirReviewRandomSearch2015}, where each of them has its own limitations and advantages. The GS is more suitable for low dimensional problems but it has difficulties to explore the finer regions in more complex scenarios, i.e., when there are few HPs to set in contexts of the large search space. The RS, on the other hand, is more flexible to explore several solutions, however, it can be computationally more expensive since it does not perform an informed search\,\cite{mantovaniRethinkingDefaultValues2021}. Meta-heuristics, such as such as \textit{Genetic Algorithms}\,(GAs)\,\cite{gomesCombiningMetalearningSearch2010}, \textit{Estimation of Distribution Algorithms}\,(EDAs)\,\cite{padiernaHyperParameterTuningSupport2017}, and \textit{Sequential Model-based Optimization}\,(SMBO)\,\cite{snoekPracticalBayesianOptimization2012} are commonly used for HPs tuning. These methods have the advantages, such as faster convergence and probabilistic nature but they may still require a large number of candidate solutions to be evaluated. However, SMBO for example has many HPs but still requires iterative evaluation of the function to be optimized. It does not eliminate the shortcoming of having to iteratively evaluate the function to be optimized\,\cite{garouthesis2022}.

\subsection{Hyperparameters tuning techniques}
A careful literature study reveals that in recent years, a variety of hyperparameters tuning methods have been employed in machine learning algorithms, as discussed in several works\,\cite{andradottirReviewRandomSearch2015,mantovaniRethinkingDefaultValues2021,gomesCombiningMetalearningSearch2010,padiernaHyperParameterTuningSupport2017}. One common approach is to iteratively build a population $\mathcal{H}$ of hyperparameter configurations and compute $\mathcal{L}(A^{(i)}, H_n, D_{train}, D_{validation})$ for each $H_n\in \mathcal{H}$. It enables the simultaneous exploration of different regions of the search space. There exist various population-based hyperparameters tuning strategies, which differ in how they update the $\mathcal{H}$ at each iteration. The following sections provide a brief description of some of these strategies.

\subsubsection{Grid Search}
The most simple method for tuning hyperparameters is the Grid Search technique. It involves exhaustive search of a subset of the hyperparameters space to select the best model from a family of models that are parameterized by a grid of values. The steps involved in this method are outlined in Algorithm,\ref{gs}.

\begin{algorithm}
	\caption{Grid Search pseudo-code. }\label{gs}
	\begin{algorithmic}[1]
		\State	$ Best_{global} \gets \texttt{NULL} $
		\For{ each $H_n \in \mathcal{H}$} 
		\State{ Sample a set of values $V=v_{i1}, v_{i2},\dots, v_{in}$ from $H_n$}
		\EndFor
		
		\For{ each $\lambda \in (H_{1},\dots, H_{k}) = (V_{1},\dots, V_{k}) $} 
		\State{$ Best_{local} \gets f(A, D, \lambda) $}
		\EndFor
		
		\State	$ Best_{global} \gets \max(Best_{local}) $
		
		\Return $Best_{global}$
	\end{algorithmic}
\end{algorithm}

Grid search is an option when dealing with a small number of hyperparameters. Nevertheless, as the number of hyperparameters increases, the computational cost required to explore the search space can become prohibitively high. Despite this limitation, the manual selection of the grid values can provide valuable insights into the behavior of the hyperparameter space surface\,\cite{bergstraRandomSearchHyperparameter2012}.

\subsubsection{Random Search}
Random Search is a method that conducts random experiments within a given search space. It can help reduce computational expenses when a large number of possible configurations are to be examined\,\cite{andradottirReviewRandomSearch2015}. Random Search usually operates iteratively on a population $P$ for a pre-defined number of iterations. In each iteration $i$, $P(i)$ is extended, or updated, by a randomly generated hyperparameter configuration $h \in \mathcal{H}$. It has been widely observed that Random Search has produced effective optimization results for Deep Learning,(DL) algorithms\,\cite{bergstraRandomSearchHyperparameter2012}. The basic workflow of Random Search is presented in Algorithm\,\ref{rs}.

 \begin{algorithm}
	\caption{Random Search pseudo-code. }\label{rs}
	\begin{algorithmic}[1]
		\State	$ t \gets 1 $
		\State	$ Best_{global} \gets \texttt{NULL} $
		\While { Stoping criteria not satisfied}
		\State{ Generate a population $P(t)$ randomly}
		
		\For{ each $p_i \in P(t)$} 
		\State{$ f_{p_i} \gets f(A, D, p_i) $}
		\EndFor
		\State{$ Best_{local} \gets \max(f_p) $}
		
		\If {$ Best_{local} \ge Best_{global} $}
		\State{$ Best_{global} \gets Best_{local} $}
		\EndIf
		
		\State{$ t \gets t+1 $}
		\EndWhile
		
		\Return $Best_{global} $

	\end{algorithmic}
\end{algorithm}

\subsubsection{Bayesian Optimization}
Grid and random search are both optimization techniques that do not consider previous evaluations, making them stateless. To overcome this limitation, Bayesian optimization can be used to automatically find optimal configurations by treating the choice of algorithm configurations as a black-box global optimization problem.

Bayesian optimization is an adaptive hyperparametric search method that predicts the most beneficial hyperparameter combination based on the tested combinations\,\cite{Sadasc}, making it a state-full technique. Let us assume that the function $f(x)$ of hyperparameter optimization follows a Gaussian process, such that $p(f(x)|x)$ is a normal distribution. The Bayesian optimization process is modeled as a Gaussian process based on the results of previous $N$ group experiments, $H=\{x_n, y_n\}_{n=1}^N$, and calculates the posterior distribution $p(f(x)|x,H)$ of $f(x)$. 
After obtaining the posterior distribution of the objective function, an acquisition function $a(x,H)$ is defined to trade off in sampling where the model predicts a high objective and sampling at locations where the prediction uncertainty is high. The goal is left to maximize the acquisition function to determine the next sampling point. The Bayesian optimization process is summarized in Algorithm\,\ref{bo}.

 \begin{algorithm}
	\caption{Bayesian Optimization pseudo-code. }\label{bo}
	\begin{algorithmic}[1]
		\State{$ H \gets \emptyset $}
		\For{$t  itteration \in \mathbb{N}$} 
		\State{$ s' \gets argmax_xa(x,H) $}
		\State{ evaluate $ y'=f(x')$}
		\State{$ H \gets H 	\cup (x', y')$}  
		\State{ Remodeling Gaussian processes according to H, calculate $p(f(x)|x,H)$}
		\EndFor
		\Return $H$

	\end{algorithmic}
\end{algorithm}

%%%

\subsubsection{Genetic Algorithms}
Genetic Algorithms\,(GA) is among the Bio-inspired techniques. It is based on natural processes and it has also been largely used for HPs tuning\,\cite{gomesCombiningMetalearningSearch2010}. In these techniques, the initial population $P$ generated randomly or according to the background knowledge, is updated in each iteration according to operators based on natural selection and evolution. The Genetic Algorithm general pseudo-code is presented in the Algorithm\,\ref{ga}.

 \begin{algorithm}
	\caption{Genetic Algorithm pseudo-code. }\label{ga}
	\begin{algorithmic}[1]
		\State	$ t \gets 0 $
		\State{ Generate initial population $P(0)$}
		\State{ Evaluate the current population $P(t)$}
		\While{ Stopping criteria not satisfied}
		\State	$ t \gets t+1 $
		\State{Select population $P(t)$ from $P(t-1)$}
		\State{Apply crossover operators in $P(t)$}
		\State{Apply mutation operators in $P(t)$}
		\For{ each new individual $i$ in the current population $P(t)$}
		\State{Evaluate individual $i$ fitness}
		\EndFor
		\EndWhile
		\State{$ Best_{global} \gets $ best individual $i$ from $P(t)$}\\
		\Return  $ Best_{global} $

	\end{algorithmic}
\end{algorithm}

The choice of an appropriate search method depends on the search space because of the underlying assumptions made by different search methods. Bayesian approaches based on Gaussian processes\,\cite{kandasamyMultifidelityGaussianProcess2019,feurerEfficientRobustAutomated2015a} and gradient-based approaches\,\cite{bergstraRandomSearchHyperparameter2012,maclaurinGradientbasedHyperparameterOptimization2015} are generally limited to continuous search space. In contrast, tree-based Bayesian\,\cite{hutterSequentialModelBasedOptimization2011, bergstraAlgorithmsHyperparameterOptimization2011}, evolutionary strategies\,\cite{olsonEvaluationTreebasedPipeline2016}, and random searches are more flexible and can be applied to any search space. The application of reinforcement learning to general hyperparameter optimization problems is restricted because of the difficulty in learning a policy over large continuous action spaces.

In this section, a formal definition of the algorithm selection and related HPs tuning problem is presented. The following sections also describe the main techniques that are often used to solve the problem. It ranges from the most straightforward techniques Grid Search, Random Search, and meta-heuristic like Genetic Algorithms to more complex approaches like Bayesian Optimization. All of these techniques can improve the predictive performance of the final induced models, but they can also be very time consuming to find suitable HPs settings. Moreover, it is not guaranteed that the tuning process will lead to either improved ML models or significant improvements. Therefore, the complete or partial automation of roles that require human skills would be welcomed with great interest. Based on this motivation, Automated Machine Learning\,\cite{hutterAutomatedMachineLearning2019} has become one of the most relevant research topics not only in the academic field but also in the industrial field\,\cite{ waringAutomatedMachineLearning2020}. The main purpose of AutoML is to provide seamless integration of ML in various industries, which can reduce the exigencies of data scientists by enabling domain experts to build ML applications automatically without extensive knowledge of statistics and machine learning. In the following section, we discuss practical approaches for automated machine learning methodologies in detail.

\section{Automated machine learning}
\label{sec:automl}

Although ML algorithms do not essentially require human interference while learning and preparing data but interference of skilled data scientists is often solicited in order to discover the right algorithm and tweaking it to get the best results. Data scientists usually try different techniques for pre-processing, which also differ for multiple ML algorithms, to eventually come up with the combination that is the most efficient\,(see Figure\,\ref{fig:humain_process}). Although, this approach is accustomed, but these processes heavily rely on human-dependent and expertise in computer science, mathematics, and statistics, as well as a deep understanding of business knowledge in the area of the data being processed\,\cite{Garouani_2022,Garouani2022-sf}. 

AutoML is an emerging research field that aims to enable non-ML experts to effectively generate and configure data analysis solutions for real-world problems without any special assistance or intervention. AutoML approach leverages human expertise by allowing users to define algorithm constraints and performance metrics, saving time and effort for knowledgeable practitioners\,\cite{vainshteinHybridApproachAutomatic2018a,10.1007/978-3-031-23615-0_33}. The central problem in AutoML is to solve a machine learning task with respect to a dataset and performance criterion. Hence, the goals of AutoML are to democratize ML, save time and effort for practitioners, and prevent methodological errors and over/under-estimation of performance\,\cite{Garouani2022-hm}.

Several approaches\,\cite{kotthoffAutoWEKAAutomaticModel2019b,olsonTPOTTreeBasedPipeline2019a} have been proposed to automate machine learning, ranging from automatic data pre-processing to automatic model selection. Some approaches aim to simultaneously select the right learning algorithm and find the optimal configuration of hyperparameters, known as the combined algorithm selection and hyperparameters optimization problem\,\cite{kotthoffAutoWEKAAutomaticModel2019b,olsonTPOTTreeBasedPipeline2019a}. A solver for this problem selects an algorithm and tunes it to achieve the highest validation performance among all possible algorithm and hyperparameter combinations.

Owing to the immense potential of AutoML, different learning paradigms have been applied to this task, and tools are available at the disposal of the research community\,\cite{bilalliPRESISTANTDataPreprocessing2018a}. These tools include but are not limited to Auto-sklearn\,\cite{feurerAutosklearnEfficientRobust2019}, AutoWEKA\,\cite{kotthoffAutoWEKAAutomaticModel2019b}, TPOT\,\cite{olsonTPOTTreeBasedPipeline2019a} as well as commercially available tools such as RapidMiner\,\cite{kotuPredictiveAnalyticsData2014}, H2O\,\cite{H2OAiAI}, Big ML\,\cite{BigML}, and Data Robot\,\cite{DataRobot}. Since 2015, a competition has been ongoing\,\cite{automlchallenges} with a focus on budget-limited tasks for supervised learning in pursuit of the goal of AutoML. 

Over the past few years, there has been considerable research and development on automating analytics workflows and their components. Two of the main approaches that have emerged are the use of ontologies and meta-learning, which aim to support data scientists and those new to the ML domain.

\subsection{Ontology based approach}
The ontology-based semantic data mining approach aims to leverage formal ontology in the process of data mining. A well-designed ontology can assist both data analysts and neophyte machine learning domain experts to select appropriate modeling techniques to build specific models and rationalize the techniques and models selected in a number of ways. This, in turn, can assist them in building specific models as well as understanding the reasoning behind the selected techniques and models. 

By expressing domain expertise in a formal structure, logical reasoning can be used to narrow down the search space and identify the most predictive model for a given problem. This allows for more efficient and effective data analysis, as well as for the discovery of previously unknown relationships and patterns within the data. Additionally, the use of formal ontologies provides a common vocabulary for communication between domain experts and machine learning practitioners. This approach allows experts to reason about the data and its properties using logical inference, narrowing down the search space and identifying the most predictive model for a given problem. The ontology-based approach for meta-learning provides a more structured and informed way of performing meta-learning, combining both domain knowledge and data-driven insights to develop more accurate and effective models, when it comes to the interpretation and application of machine learning models.

The ontology-based approach for meta-learning differs from the traditional data-driven approach in that it combines knowledge of the data-mining process with a formal ontology and knowledge base. Unlike the data-driven approach that relies solely on statistical analysis of data to identify patterns and relationships between different tasks, the ontology-based approach utilizes a pre-defined set of concepts, relationships, and rules to represent domain knowledge and expertise in a formal structure.

An ontology-based Intelligent Discovery Assistant\,(IDA) described by Bernstein \textit{et al.}\cite{bernsteinIntelligentAssistanceData2002} analyzes an input dataset to extract meta-data and generates all possible workflows from the ontology that are valid within the characteristics of the input.
The system sorts the recommendations based on user-specified criteria such as simplicity, performance metric, etc. Similarly, Nural \textit{et al.}\cite{mustafavnuralOntologybasedSemanticsVs2017} compared the meta-learning approach to the ScalaTion ontology-based\,\cite{nuralUsingSemanticsPredictive2015} suggestion approach on 114 datasets\,\cite{nuralUsingMetalearningModel2017a}. Using ScalaTion, each dataset is provided as an input to the suggestion engine and the suggested modeling technique is recorded. The ontology-based approach achieved an accuracy of 51\% compared to 75\% with meta-learning while predicting the top-1 performing technique.

\subsection{Meta-learning based approach}
\label{mtl_sec}

When acquiring new skills, individuals seldom start from scratch; instead, they draw on previously acquired skills from related tasks, reuse effective strategies, and concentrate on approaches that have worked well in the past,\cite{lakeBuildingMachinesThat2016}. As individuals acquire more skills, learning new ones becomes simpler, requiring fewer examples and less trial and error. In essence, they develop the ability to learn across tasks. Similarly, when constructing machine learning models for a particular task, we frequently rely on our understanding of related tasks or our implicit knowledge of how ML techniques behave to make informed decisions.

Several ML algorithms have been proposed for prediction tasks. However, since each algorithm has its inductive bias, some of them can be more appropriate for a particular dataset. A ML algorithm may produce a higher predictive performance on a data set if that algorithm has more appropriate bias for the given dataset. Thus, non-experienced users become overwhelmed and require support\,(e.g., to recommend the algorithm along with the related HPs configuration). In this context, the recommendation of the most adequate ML algorithm configuration for a new dataset is investigated in a research area known as Meta-Learning\,(MtL).

Meta-learning or learning to learn is the process about learning how machine learning algorithms perform across a range of tasks\,\cite{Garouani2022}. It aims to learn\,
\begin{itemize}
    \item  which algorithm could work well for a dataset with certain characteristics, or 
    \item which hyperparameters can give a good performance for given dataset to achieve the given tasks, or
    \item which dataset can efficiently achieve the task for a given algorithm, or
    \item which task can be achieved efficiently by an algorithm on a given dataset. 
\end{itemize}

Thus, the MtL approach examines the relationship between problem domains and learning strategies. The primary goal is to determine the most effective algorithm for a particular task and seeks to understand when a specific learning strategy is more appropriate than others. This task is also commonly referred to as “Algorithms Selection”\,\cite{vilaltaUsingMetaLearningSupport2004}. Meta-learning for algorithm selection and hyperparameter tuning is based on the assumption that\,: “\textit{Algorithms show similar performance for the same configuration for similar problems}”. It aims to create a meta-model that maps problem characteristics to algorithm performance, which can be used to solve these problems.

\begin{figure}[h!]
	\centering
	\resizebox{1\hsize}{!}{\includegraphics{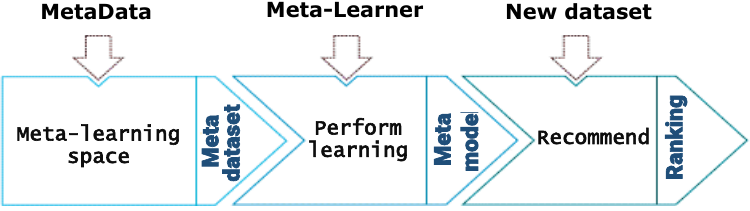}}
	\caption{The meta-learning process.}
	\label{fig:mtl_process}       % Give a unique label
\end{figure}

Meta-learning involves systematically and data-driven learning from previous experiences, and it typically consists of three main phases, as depicted in Figure \ref{fig:mtl_process}. In the first phase, a meta-learning space is created using \textit{meta-data} that describes past learning tasks and models. This meta-data includes \textit{meta-features} that describe the dataset's characteristics and \textit{meta-responses} that provide a performance measure for data-mining algorithms on those datasets. In the second phase, a predictive \textit{meta-model} is developed from the meta-dataset created in the first phase, which extracts and transfers knowledge that guides the search for optimal models for new tasks. Finally, in the third phase, when a new dataset is encountered, its characteristics are extracted, and the predictive \textit{meta-model} is used to recommend the most promising ML algorithms with their corresponding hyperparameter configurations. The key challenge in meta-learning is to learn from previous experiences in a systematic and data-driven manner to effectively guide the search for optimal models for new tasks.

%%% à faire: correction dans la figure (Meta Mode -> Meta Model) (MetaData-> Meta-Data

\begin{figure*}[b]
	\centering
	\includegraphics{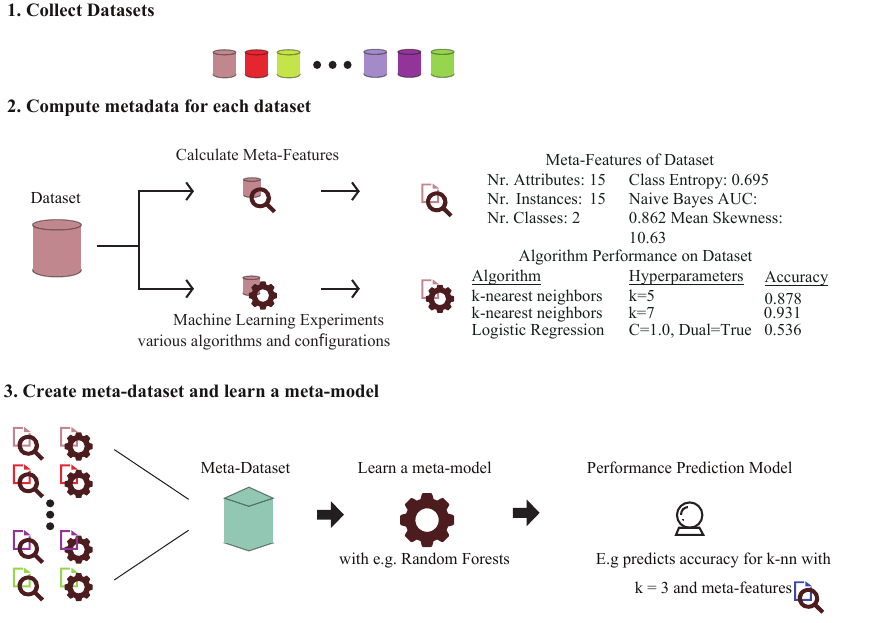}
	\caption{An overview of the steps to create a meta-model.}
	\label{fig:mtmodel_process} 
\end{figure*}

Figure,\ref{fig:mtmodel_process} illustrates the procedure to build a \textit{meta-model}. It is necessary to have a \textit{meta-dataset} that can be used to learn knowledge in order to train a meta-model. This dataset contains details about machine learning experiments, including the employed algorithm, the used hyperparameter configuration, and the performance of the resulting model. Additionally, for each experiment, the meta-dataset includes information about the used dataset, which is conveyed through \textit{meta-features} such as the number of attributes, classes, kurtosis, skewness, and other relevant information.

The initial step involves selecting a group of datasets to be utilized in machine learning experiments, as depicted in the first stage of Figure,\ref{fig:mtmodel_process} to generate a meta-dataset. In the second stage, two tasks must be performed for each dataset. In this respect the meta-features which are broadly speaking the characteristics of the dataset must be computed at the initial stage. These meta-features describe the dataset in a variety of ways, such as the number of classes, the average skewness of numerical features, and the accuracy of basic classifiers, among other factors. More information regarding the various meta-features is presented in the subsequent section. Secondly, machine learning experiments must be conducted on the datasets. The type of experiments should be consistent with the kind that the meta-model. Finally, the outcomes from stage two are combined to form a comprehensive meta-dataset. This dataset includes a row for every machine learning experiment and describes the employed algorithms and associated hyperparameter configurations, meta-features of the dataset on which the experiments are performed, and the performance results.

This methodology is applied to further specify the objectives for the later on following the establishment of the problem statement. For this purpose, we introduce an optimization strategy that seeks to learn the correlations between datasets features and data mining algorithms,\cite{Brazdil2009}; instead of employing arbitrary groups of pipelines and fine-tuning relevant hyperparameters through techniques like Genetic Algorithm or Bayesian Optimization. As it can be observed that a predictive meta-model can forecast the effectiveness of a specific combination of learning algorithms in the set \textbf{$A$} and a chosen set of hyperparameters in \textbf{$H$} on a dataset \textit{$D$}, using the characteristics of a dataset.

%%% ref to Definition 1 & 2 and Algorithm 3 and 4 ?

%For instance, in a classification task, meta-learning can be used to predict the accuracy of a classification algorithm on a given dataset and hence provide user support in the mining step.

%The process of ranking consists of three phases\,(cf.Figure\,\ref{fig:mtl_process}, Chapter\,\ref{chap:automl}). 

The procedure, establishes a \textit{meta-learning space} using meta-data to initiate the learning. The meta-data contains both the characteristics of datasets and the performance metrics of data mining algorithms on those particular datasets. Later on, during the \textit{meta-learning phase}, a predictive meta-model is created, which defines the scope of competence of the data mining algorithms,\cite{besimbilalliLearningImpactData2018}. Eventually, when a fresh dataset arrives, we extract its features and input them into the predictive meta-model. Based on this, the model recommends the machine learning pipelines that have the potential to perform well on the given dataset. Consequently, in this recommendation phase, we can rank the pipelines based on their predicted impact on the given dataset by comparing the predictions obtained for the different pipelines on similar tasks.

After receiving these recommendations, a suitable meta-dataset is created and a meta-model is induced, as a result of the MtL process. The induced meta-model represents a mapping between the meta-features that describe the datasets and the predictive performance achieved by the group of learning algorithms when applied to the concerned datasets. Therefore, the quality of the meta-features is crucial for the predictive performance of the meta-models. As a result, the meta-models can potentially recommend the best algorithms for a new, unseen problem.

The two fundamental concepts of meta-learning are the \textit{meta-data} and the \textit{meta-features} which provide a more detailed structure of the \textit{meta-model}. In the following, these two concepts are briefly discussed.

\subsubsection{Meta-data}
The meta-data for learning is essential to develop a useful meta-model. Meta-data in this regard, includes the critical information necessary for creating a meta-dataset. It includes both \textit{meta-features}, which are the predictors, and the algorithm's performance measures, which are the \textit{meta-responses}.

%%% meta-model ? see line 634
\paragraph{Meta-features}
An essential question in MtL is \textit{how to obtain relevant information for characterizing specific tasks?} Researchers have attempted to address this question by examining dataset properties that impact the performance of learning algorithms, measuring performance directly\,\cite{pfahringerMetaLearningLandmarkingVarious2001}, exploring alternatives\,\cite{meskhiLearningAbstractTask2021}, and developing or adapting new measures based on existing ones\,\cite{garouani_iceis23}. Various types of meta-features have been created, ranging from simple ones like the number of samples in a dataset to more complex ones. In this paper, we present a brief overview of the most commonly used meta-features and explain why they are indicative of model performance. Where possible, we include formulas for computing them. More comprehensive surveys can be found in the literature as cited in\,\cite{rivolliCharacterizingClassificationDatasets2019, reifAutomaticClassifierSelection2014}.

\begin{enumerate}
	\item \textbf{Simple meta-features}\\
	The basic information about a dataset is represented by simple measures, which can be directly extracted from the data. These measures are the simplest to define and have the lowest computational cost\,\cite{castielloMetadataCharacterizationInput2005,  reifAutomaticClassifierSelection2014}. They are deterministic and free of hyperparameters. In Table\,\ref{tab:SimpleMF}, we present some examples of such measures. They represent concepts related to the number of predictive attributes, target classes, instances, and missing values. These measures are crucial for characterizing the primary aspects of a dataset, and they provide valuable information that can assist in selecting an appropriate learning algorithm for a specific task.

\begin{table}[h]
	\caption{Simple measures and their characteristics.}
	\label{tab:SimpleMF}  
	\centering
	\begin{tabular}{lcl}
		\hline
		\noalign{\smallskip}
		Name	&  Formula & Rationale	\\
		\noalign{\smallskip}\hline\noalign{\smallskip}
		
		Nr instances	&  	$n$	&	Speed, Scalability\\
		Nr attributes		&  	$p$	&	Dimensionality\\
		Nr classes		&  	$c$	&	Complexity, imbalance\\
		Nr missing values&  $m$	& 	Imputation effect\\
		Nr outliers		&   $o$ &	Data noisiness\\
		attrToInst		&       &	Dimensionality  \\
		instToAttr		&      &	Sparsity  \\
		
		\noalign{\smallskip}\hline                                              
	\end{tabular}
\end{table}

Information about the number of instances and classes are significant means to indicate the dataset size and diversity of labels; however, this information is not sufficient. A combination of this information with the number of attributes can capture additional concepts that are generally different and straightforward. 
The attrToInst and instToAttr measures can be used to capture the dimensionality and sparsity of the data, respectively. If instToAttr is too small, it risks overfitting, where the learning model considers irrelevant details in the training data, resulting in poor generalization\,\cite{kubaExploitingSamplingMetalearning2002}. Furthermore, some metric measures can evaluate the quality of the dataset, such as the number of missing values in the dataset attributes and instances as well as the total count. Because some ML algorithms can handle missing values, these measures provide crucial information for selecting appropriate algorithms.

\item \textbf{Statistical meta-features}\\
Statistical measures can be employed to extract information about the performance of statistical algorithms or data distribution, such as central tendency and dispersion\,\cite{castielloMetadataCharacterizationInput2005}. They are the most extensive and varied group of meta-features, as depicted in Table\,\ref{tab:StatMF}. These measures are deterministic and are suitable for supporting only numerical attributes.

\begin{table}[h!]
	\caption{Statistical measures and their characteristics.}
	\label{tab:StatMF}  
	\centering
	\begin{tabular}{lcl}
		\hline
		\noalign{\smallskip}
		Name	&  Formula & Rationale 	\\
		\noalign{\smallskip}\hline\noalign{\smallskip}
		
		Skewness	&	&Feature normality  	\\
		Kurtosis	&	&Feature normality	\\
		Correlation&	&Feature interdependence	\\
		Covariance&		&Feature interdependence	\\
		Sparsity&		&Degree of discreteness	\\
		Gravity&		&Inter-class dispersion 	\\
		ANOVA p-value&	&Feature redundancy  	\\
		\noalign{\smallskip}\hline                                              
	\end{tabular}
\end{table}

Correlation and covariance measures are used to capture the interdependence of the predictive attributes\,\cite{castielloMetadataCharacterizationInput2005}. Both measures are calculated for every pair of attributes in the dataset. Correlation is a normalized version of covariance. The absolute values of both statistical measures are often used, changing their range to [-1; 1] and ]-$\infty$;+$\infty$[, respectively. High values indicate a strong correlation between the attributes, which can imply redundancy in the data\,\cite{kalousisModelSelectionMetalearning2000}.

%	A specific measure to capture the normality of the attributes is the nrNorm, which computes the number of attributes normally distributed. Similarly, nrOutliers counts the number of attributes that contain outliers. Normality and outliers may affect the behavior of learning algorithms, which make these measures useful in a MtL scenario.
	
\item \textbf{Information-Theoretic meta-features}\\
The information-theoretic meta-features aim to capture the amount of information present in the data. Table,\ref{tab:InfoMF} illustrates the information-theoretic measures, which usually require categorical attributes and are mostly used for representing classification problems. These measures are directly computed, free of hyperparameters, deterministic, and robust. In terms of their semantic meaning, they describe the variability and redundancy of the predictive attributes that represent the classes.

\begin{table}[h!]
	\caption{Information-theoretic meta-features and their characteristics.}
	\label{tab:InfoMF}  
	\centering
	\begin{tabular}{lcl}
		\hline
		\noalign{\smallskip}
		Name	&  Formula & Rationale 	\\
		\noalign{\smallskip}\hline\noalign{\smallskip}
		
		Class entropy &	$H(C)$	&Class imbalance \\
		
		Norm. entropy &$\frac{H(X)}{log_2n}$&	Feature informativeness \\
		Mutual inform. &$MI(C, X)$&	Feature importance\\
	
		Uncertainty coeff.&	$\frac{MI(C,X)}{H(C)}$&	Feature importance\\
		
		Equiv. nr. feats  & $\frac{H(C)}{\overline{MI(C,X)}}	$&Intrinsic dimensionality \\
		
		Noise-signal ratio &  $\frac{\overline{H(X)}-\overline{MI(C,X)} } {\overline{MI(C,X)}}$&	Noisiness of data\\
				 
		\noalign{\smallskip}\hline                                              
	\end{tabular}
\end{table}

	According to \cite{rivolliCharacterizingClassificationDatasets2019}, the entropy of both the predictive attributes and target values measures the average level of uncertainty present in the predictive and class attributes, respectively. The entropy of the predictive attributes assesses all of the predictive attributes, making it possible to obtain an overview of their ability to distinguish between classes. Conversely, the entropy of the target values measures the amount of information necessary to specify a single class. From a machine learning perspective, a predictive attribute with low entropy is less effective at distinguishing between classes,\cite{castielloMetadataCharacterizationInput2005}, while a target attribute with low entropy indicates a high level of class purity.

	In prediction tasks, it is typical to designate a variable or several variables as the response. Additionally, meta-features that evaluate the relationship between the predictors and the used response. These measures fall into two categories: Landmarking and Model-based\,\cite{gijsbersAutomaticConstructionMachine2017}. Yet, when applied to larger datasets, the computational expense associated with these measures can be considerable.

	\item \textbf{Landmarking}\\	
Landmarking is an approach for characterizing datasets based on the performance of a set of simple and fast learners that extract information from the learning models. Examples of such learners include decision stumps, naive Bayes classifiers, and linear discriminant analysis,\cite{gijsbersAutomaticConstructionMachine2017}. The most commonly used landmarking measures are listed in Table \ref{tab:LandMF}. To be considered a good landmarker, the runtime complexity of the method should be at most $O(n\,log(n))$. Additionally, when employing multiple landmarkers, they should have different biases,\cite{gijsbersAutomaticConstructionMachine2017}. If two distinct landmarkers display similar performance across all datasets, it is likely sufficient to use only one of them,\cite{furnkranzEvaluationLandmarkingVariants2001}. These measures have been studied in various works, including \cite{reifAutomaticClassifierSelection2014,feurerAutosklearnEfficientRobust2019,feurerEfficientRobustAutomated2015a}.

\begin{table}[h!]
	\caption{Landmarking meta-features and their characteristics.}
	\label{tab:LandMF}  
	\centering
	\begin{tabular}{lcl}
		\hline
		\noalign{\smallskip}
		Name	&  Formula & Rationale 	\\
		\noalign{\smallskip}\hline\noalign{\smallskip}
		
		Landmarker(\texttt{1NN}) &$ P(\theta_{1NN}, t_j)$ &	Data sparsity \\
		Landmarker(\texttt{Tree}) &$ P(\theta_{Tree}, t_j)$ &	Data separability  \\
		Landmarker(\texttt{Lin})  &$P(\theta_{Lin}, t_j)$ &	Linear separability  \\
		Landmarker(\texttt{NB}) &$ P(\theta_{NB}, t_j)$  &	Feature independence    \\
		Relative LM  & $P_{a,j} - P_{b,j} $&	Probing performance   \\
		Subsample LM   &$P(\theta_i, t_j , s_t)$  &Probing performance   \\

		\noalign{\smallskip}\hline                                              
	\end{tabular}
\end{table}

 \item \textbf{Data complexity }\\
The measures known as complexity measures are proposed in\,\cite{garciaNoiseDetectionMetalearning2016} with the aim of evaluating the difficulty of classification problems by taking into account attributes overlap, class separability, and geometric/topological properties. Table \ref{tab:DCMF} provides an overview of the key features of these measures. A more comprehensive list of complexity measures is available in\,\cite{lorenaDataComplexityMetafeatures2018, hoComplexityMeasuresSupervised2002a}.

\begin{table}[h!]
	\caption{Data complexity meta-features and their characteristics.}
	\label{tab:DCMF}  
	\centering
	\begin{tabular}{lcl}
		\hline
		\noalign{\smallskip}
		Name	&  Formula & Rationale 	\\
		\noalign{\smallskip}\hline\noalign{\smallskip}
		
		Fisher’s discrimin&$\frac{(\mu{c1} - \mu{c2})^2}{ \sigma_{c1}^2 - \sigma_{c2}^2  }$&Separability classes $c_1, c_2$ \\
		Volume of overlap & &Class distribution overlap		\\
		Concept variation& &Task complexity 	 \\
		Data consistency & &Data quality 		 \\

		\noalign{\smallskip}\hline                                              
	\end{tabular}
\end{table}

\item \textbf{Model structure-based meta-features}\\
%Similar to landmarking, model-based features are constructed by running machine learning algorithms on the dataset. The meta-features of this group are information extracted from a predictive learning model, in particular, a Decision Tree\,(DT) model. They characterize a dataset by how complex is the model induced, which, for DT, can be the number of leaves, the number of nodes and the shape of the nodes tree\,\cite{bensusanHigherorderApproachMetalearning2000, pengDecisionTreeBasedData2002, reifAutomaticClassifierSelection2014}.
Unlike the previous data characterization methods that rely on the data distribution, the model-based approach is an indirect characterization method that utilizes decision tree models to extract information about the hidden structures of the data\,\cite{reifAutomaticClassifierSelection2014}. The properties of the decision tree, such as its depth and number of nodes, are used as meta-features. This approach has the advantage of considering the representation of the data set in a special structure for extracting information about the learning complexity, rather than relying solely on the distribution of the data. However, it comes with a relatively high computational cost. Table,\ref{tab:ModelMF} summarizes the decision tree model meta-features.

\begin{table}[h!]
	\caption{Model-based meta-features and their characteristics.}
	\label{tab:ModelMF}  
	\centering
	\begin{tabular}{lcl}
		\hline
		\noalign{\smallskip}
		Name	&  Formula & Rationale	\\
		\noalign{\smallskip}\hline\noalign{\smallskip}
		
		Nr nodes, leaves, and branches&$ |\eta|, |\psi|$&Concept complexity\\
		Nodes per feature &$|\mu\,X|$&Concept complexity \\
		Leaves per class&$\frac{|\psi_c|}{|\psi|}$&Feature importance 	\\
		Leaves agreement &$\frac{n_{\psi_i}}{n}$&Class complexity 	\\
		Information gain&&Class separability \\
	
		\noalign{\smallskip}\hline                                              
	\end{tabular}
\end{table}
\end{enumerate}
Several non-traditional characterization measures have been proposed in the literature, although they are not commonly used in MtL studies because of their high computational complexity or domain bias. However, they may be useful in certain learning scenarios and MtL problems, as some studies have shown promising results with these measures\,\cite{garciaNoiseDetectionMetalearning2016}.

\subsubsection{Meta-target}

%\subsubsection{Meta-model}

Once the meta-dataset with all necessary metadata has been generated, the objective is to develop a predictive meta-model capable of learning the complex relationships between the characteristics of the tasks, i.e., meta-features, and the suitability of a specific ML pipeline. This can help recommend the most effective ML algorithm configuration, $A^{(i)*}$, given the meta-features $\mathcal{M}$ of the new task $t_{new}$.

Each task, $t_j\in \mathcal{T}$, can be described by a vector, $m(t_j)=(m_{j,1},...,m_{j,K})$, of $K$ meta-features. This allows for the definition of a task similarity measure based on, for instance, the Euclidean distance between $m(t_i)$ and $m(t_j)$, so that information can be transferred from the most similar tasks to the new task $t_{new}$.

Different meta-learners have been employed in the literature, such as k-nearest neighbors\,(KNN), decision trees, and XGBoost,\cite{besimbilalliLearningImpactData2018}. Various approaches have investigated the use of MtL to provide recommendations on the ML pipeline to use, with suggestions taking one of the following forms, depending on the aspect to which MtL is applied:

\begin{description}
	\item[a) List of applicable algorithms] \hfill \\
	This category refers to approaches where a single algorithm or a set of algorithms are suggested to perform best on the given dataset, based on a performance criterion. The approaches predict the settings of HPs without actually evaluating the model on the new dataset\,\cite{vilaltaUsingMetaLearningSupport2004}. For example, \cite{zhongguoChoosingClassificationAlgorithms2017} employed MtL to recommend the optimal algorithm for a given task. They evaluated C4.5, kNN, and SVM classifiers with some of their HPs using a grid design over 100 UCI\,\footnote{\url{http://archive.ics.uci.edu/ml/index.php}} datasets, measuring performance using AUC. The kNN algorithm was used as a meta-model to recommend the best HPs setting for new datasets. Similarly, \cite{reifAutomaticClassifierSelection2014} evaluated different ML algorithms on 54 datasets and used performance predictions to develop a MtL system for automatic algorithm selection. In\,\cite{priyaUsingGeneticAlgorithms2012}, the authors conducted a similar study with SVMs, but optimized HPs and performed feature selection using Genetic Algorithm on six meta-learners,(kNN, SVM, J48, JRip, NB, and Bagging) over 78 classification datasets. They assessed HPs settings using a 5-fold cross-validation strategy and the Mean Absolute Deviation evaluation measure.
	
	Meta-models can also generate a ranking of the top-K most promising configurations. For example, in \cite{pintoAutoBaggingLearningRank2017}, the "autoBagging" tool is proposed, an AutoML system that automatically ranks Bagging workflows considering four different Bagging HPs. The authors explore past performances and datasets characterization and evaluate the results at the meta-level using a Mean Average Precision\,(MAP) measure in a Leave-One-Out Cross-Validation\,(LOO-CV) strategy. They conduct experiments on 140 OpenML\,\footnote{\url{https://www.openml.org/}} datasets and 146 meta-features extracted with post-processing aggregation functions. XGBoost is used as a meta-learner to predict the ranking of workflows. A ranking provides an ordered set of suggestions to try, and users may have preferences about the algorithm used based on, for example, computational efficiency or model interpretability.

	\item[b) Predict HPs tuning necessity and training runtime] \hfill \\ 
	    An alternative method to assess the need for hyperparameter tuning and training runtime is through the use of approach presented in\,\cite{molinaMetalearningApproachAutomatic2012}. It utilizes MtL to determine the necessary HPs tuning for the C4.5 algorithm by assessing 14 academic datasets using five simple meta-features. Their approach predicts whether different HPs settings will increase, decrease or maintain predictive performance of induced DTs. Similarly,\,\cite{riddUsingMetalearningPredict2014} use MtL to identify when HPs tuning leads to significant accuracy improvement using a PSO technique\,\cite{Marini_2015} in 326 binary classification datasets. However, no statistical analysis is performed instead the authors use different thresholds to determine the improvement.
	    
	    In contrast, a meta-regressor can be trained to predict the run-time of algorithms during training or prediction phases based on different HPs settings. Reif \textit{et al}.\,\cite{reifMetalearningEvolutionaryParameter2012} predicted the training run-time of kNN, SVM, MLP, and DT classifiers by evaluating discrete HPs settings on 123 classification datasets. They used Pearson Product Moment Correlation Coefficient\,(PMCC) and Normalized Absolute Error\,(NAE) as performance measures for HPs assessment. Similarly, \cite{yangOBOECollaborativeFiltering2019} predict predictive performance and runtime using polynomial regression based on the number of instances and features.
	
	\item[c) Estimate predictive performance for a given HPs setting] \hfill \\
	    Meta-models have the capability to predict the performance of a given configuration on a specific task based on its meta-features. The goal is to estimate the performance of the configuration and this task is approached as a regression problem. This enables an assessment of whether the intended configuration is worthwhile to evaluate in an optimization procedure. Early studies utilized linear regression and rule-based regressors to predict the performance of a discrete set of configurations and then rank them accordingly,\cite{bramerEstimatingPredictiveAccuracy2013}.

        To predict the performance of a set of ML algorithms, \cite{guerraPredictingPerformanceLearning2008} trained an SVM meta-regressor for each classification algorithm. This model predicts the accuracy, under default settings, of the classification algorithm on a new task  based on its meta-features. Similarly, \cite{reifAutomaticClassifierSelection2014} developed a meta-regressor trained on more meta-data to predict its optimized performance.

        Recently, \cite{wistubaSequentialModelFreeHyperparameter2015} adapted the acquisition function of surrogate models by an optimized meta-model. They evaluated several hyperparameters in a holdout procedure across 105 datasets and leveraged the meta-knowledge to predict the performance of new hyperparameter settings for new datasets. The base-level algorithms explored were AdaBoost and SVM.
	
\end{description}

\noindent
%Table\,\ref{tab:Summary} present a comprehensive list of studies that either embedded or used MtL to cope with these tasks.

\,

Research papers from the detailed literature survey that either embedded or used MtL to cope with these tasks are summarized in Table 1 in the supplemental file. For each work, it is shown the task or meta-target, the learner and meta-learner used, the number of datasets that have been considered along with the meta-features used to characterize those datasets. Related surveys are summarized afterward.

%Summary of Literature Overview
%\pagebreak
\subsection{Summary of literature overview}
\label{Summary_sec}

The literature review on MtL related works, leads to identify some interesting aspects\,\cite{Garouani_2023}. Overall, the following aspects can be observed\,:

\begin{itemize}
	\item  most of the studies create the meta-data using the straightforward grid Search technique to tune the algorithms;

	\item  most of them evaluate the resulting models with a holdout or single cross-validation re-sampling procedure and the simple Accuracy evaluation measure;

	\item  Most studies have utilized fewer than 100 datasets, with some exceptions that have utilized more than 100 datasets\,\cite{pintoAutoBaggingLearningRank2017,wangFeatureSubsetSelection2013,reifPredictionClassifierTraining2011}. However, all of these studies are limited to binary classification problems. There is no clear consensus in the literature regarding the optimal number of datasets for meta-learning. However, it is essential to use a sufficient number of datasets to effectively map the feature space into the performance space\,\cite{khan_2020, munozInstanceSpacesMachine2018};

 	\item  	at the meta-level, most of the studies considered instance based meta-learners\,(kNN);

	\item  different algorithms are tried at the meta-learning level but mostly concentrated by the same study and consider just one evaluation measure;

	\item  Most of the models investigate a small number of categories\,(simple and statistical) meta-features to characterize the datasets. Therefore, different approaches to generate meta-features are not well explored in the literature;

	\item  few of them provide the complete resources for the reproducibility of experiments;

	\item an end-to-end MtL resolver for the CASH problem has neither been proposed nor investigated;

	\item  none of the studies found by the author combined all these previous issues or limitations. 
	
\end{itemize}

The research community can acquire valuable meta-knowledge for future work by investigating various experimental configurations, meta-features, and methodologies applied to the learning tasks. Therefore, exploring these aspects can open up new horizons in the field of meta-learning.

\subsection{Related surveys}

Meta-learning based algorithm selection has been used in a variety of domains, such as automated machine learning\,\cite{garouaniUsingMetalearningAutomated2022}, natural language processing\,\cite{Garouani2022-hm}, computer vision\,\cite{Ghareh_Mohammadi_2019}, bio-informatics\,\cite{Arredondo_2015} and many others. These applications are mainly focused on finding the best performing model for a given data set or task. Hence enormous amount of literature is available in this regard. In this part, we summarize and discuss previous related surveys which are linked to our work.

To begin with, a survey paper by Smith-Miles, K. A.\,(2008)\,\cite{Smith_Miles_2009} provided a cross-disciplinary perspectives on Meta-Learning. The authors begin by providing background information about what defines a successful approach to this type of automated machine learning technique before describing different categories and architectures within it. Among others, metric based approaches, optimization based models and memory augmented networks are more notable. The authors, then discuss in detail the each one’s advantages/disadvantages before discussing potential applications across a variety fields from bio-informatics, cryptography, and other fields. 

A more recent survey is provided by Huimin Peng\,(2020)\,\cite{2004.11149} which provides an overview of the current state of meta-learning, including its theoretical foundations and application domains. The author begins with a discussion of the various types of meta-learning algorithms, as well as their theoretical properties and practical limitations. Later on, several recent trends in applications are highlighted that have emerged from research on meta-learning, such as transfer learning, lifelong learning, multi-task learning, few shot and one shot learning. Finally, a note is given on open challenges in the field and the future directions for research on meta-learning.

In 2020, Irfan Khan \textit{et al}.\,\cite{Khan_2020}, presented a literature survey which focuses on meta-learning for algorithms selection in particular. This paper looks into the different types of meta-learners available along with their application to various domains such as optimization problems and data mining tasks. This comprehensive survey paper covers the recent developments in meta-learning, including a definition of meta-learning, its applications to various tasks, algorithms used for meta-learning, challenges to be addressed in future research. Afterward, they present a comparative evaluation of some of the prominent methods for classifier selection based on 17 classification algorithms and 84 benchmark datasets.

In comparison, our survey offers an up-to-date, comprehensive review of the research done in the MtL based automated machine learning field. We attempt to provide better understanding of the algorithms selection and configuration problems. It give insight details on the current trends and highlights key findings from recent studies. Additionally, we provide a comparison of different approaches and methods used by researchers in the field. 
%By providing such a comprehensive study at existing work, our survey encourages further research into this area and may even lead to new insights or discoveries that can improve our understanding of the subject matter.

We propose a large-scale open source knowledge base of more than \textit{4 millions} previous learning experiences for developing meta-learning based algorithms selection systems which is largely missing in the literature. It provides a comprehensive set of data for developing meta learning based algorithm selection systems. The knowledge base contains real-world datasets, algorithms and their performance results on the datasets, which could help to reduce the time needed for training and testing different algorithms. Furthermore, this knowledge base could provide new insights into the performance of different algorithms in various domains and tasks. This may enable researchers to build more reliable and accurate algorithm selection systems that are tailored to specific application domains.

\subsection{Background on AutoML Systems}
AutoML has become a popular research topic, with many different approaches and accompanying open-source software available. Bayesian optimization, a technique based on Bayes' theorem, is commonly used for hyperparameter optimization in ML algorithms\,\cite{feurerEfficientRobustAutomated2015a, thorntonAutoWEKACombinedSelection2013}. Other simpler techniques, such as grid search and random search, are also used. Meta-Learning is another approach to hyperparameter optimization, where the AutoML system learns from its own experience of applying machine learning. A review of the literature reveals a variety of AutoML tools and platforms, with some being open-source and others commercial. Table\,\ref{tab:AutoML_SOTA} provides a comparison of some of the most popular AutoML platforms, based on factors such as cost, coding requirements, processing location, input data requirements, and supported Operating Systems.

\begin{table*}[h!]
	\centering

	\caption{Summary of related AutoML systems.}
	\label{tab:AutoML_SOTA}
	\small
	\begin{threeparttable}
		%\resizebox{\textwidth}{!}{%
		\begin{tabular}{lK{1.3cm}K{1.5cm}K{2cm}ccc}
			\hline
			\noalign{\smallskip}
			\multirow{2}{*}{System}&\multirow{2}{*}{Cost}&\multirow{2}{*}{Coding}& \multirow{2}{*}{Data type}  & \multicolumn{3}{c}{Operating system}  \\
			\cmidrule(r){5-7}     && need  &                 &Linux&	Mac &	Windows \\
			\noalign{\smallskip}\hline\noalign{\smallskip}
			
			Google\,AutoML\,\cite{AutoMLTables}&	Billable &No&		Img, Txt Tabular &\multicolumn{3}{c}{Cloud computing}	 \\
			H2O.ai\,\cite{H2OAiAI}&	Billable&	No&	Img, Txt Tabular &		\multicolumn{3}{c}{Cloud computing}\\
			Rapidminer\,\cite{kotuPredictiveAnalyticsData2014}&	Billable&	No &		Img, Txt Tabular&	Yes&Yes&Yes \\
			Auto Keras\,\cite{jinAutoKerasEfficientNeural2019a} &	Free&	No&		Image&	Yes&	Yes&	Yes \\
			Auto-Sklearn\,\cite{feurerAutosklearnEfficientRobust2019}&	Free&	Yes&		Tabular&	Yes&	No&	No \\
			ATM\,\cite{8257923}&	Free&	Yes&		Tabular&	Yes&	-&	- \\
			TPOT\,\cite{olsonEvaluationTreebasedPipeline2016}&	Free&	Yes&		Tabular&	Yes&	Yes&	Yes \\
			Auto-WEKA\,\cite{kotthoffAutoWEKAAutomaticModel2019b}&	Free&	Yes&		Tabular&	Yes&	Yes&	Yes \\

			\noalign{\smallskip}\hline                                        &                             &                                 &
		\end{tabular}%}%
	\end{threeparttable}
\end{table*}

Auto-WEKA\,\cite{kotthoffAutoWEKAAutomaticModel2019b} and Auto-Sklearn\,\cite{feurerAutosklearnEfficientRobust2019} are two popular AutoML frameworks for building machine learning pipelines\,\cite{kotthoffAutoWEKAAutomaticModel2019b}. Auto-WEKA is based on the Weka ML library\,\cite{ianh.DataMiningPractical2016} and applies Bayesian optimization to address the CASH problem. On the other hand, Auto-Sklearn is implemented on top of Scikit-Learn\footnote{\url{https://scikit-learn.org}} and extends the idea of configuring a general machine learning framework with global optimization. Auto-Sklearn also builds an ensemble of all models tested during the global optimization process to improve generalization. TPOT\,\cite{olsonEvaluationTreebasedPipeline2016} is another AutoML tool that employs genetic programming algorithms to optimize classification and regression ML pipelines. However, TPOT can only handle categorical parameters and many of the randomly assembled candidate pipelines evaluated end up as invalid, thus wasting valuable time.

Among the big market actors, Google Cloud Platform has recently launched AutoML Tables\,\cite{AutoMLTables}, which is a supervised learning service that automates the end-to-end ML process, but it is only available on Google Cloud as a managed service of the commercial framework. Despite providing partial or complete ML process automation, each of these tools has its unique advantages and disadvantages, as they work differently and target different dataset structures, platforms, algorithms, or end-users. For example, Auto-Sklearn is integrated with Python, but it only supports structured data and \textit{Linux Operating System}. Auto-WEKA is a tool that comes with a graphical user interface\,(GUI) and supports Weka machine learning algorithms. However, it has limitations as it is only capable of handling statistical algorithms. RapidMiner, on the other hand, is a tool that allows feature engineering, but its usage requires expert guidance. Google AutoML is a powerful tool that supports most datasets and algorithms; however, it is cloud-based, and its usage is mostly commercial for dedicated data processing.

Auto-WEKA provides a user-friendly interface that can be useful for researchers and practitioners who are new to the field of ML. However, due to its limitation to statistical algorithms, it may not be the best tool for researchers who require more advanced ML algorithms. RapidMiner is a flexible tool that allows for feature engineering, but its usage requires expert guidance, which can be a limitation for researchers who are not experienced in the field. Google AutoML is a powerful tool that supports most datasets and algorithms, making it a preferred choice for researchers who require advanced ML algorithms. However, its usage is mostly commercial, which can be a limitation for researchers with limited budgets.

It is important for researchers to consider the limitations and benefits of these tools when selecting the appropriate one for their research needs. By exploring and comparing different experimental setups, meta-features, and procedures applied to the learning tasks, researchers can gain valuable insights and open up new horizons in the field of machine learning.

\section{Benchmark Knowledge base}
\label{sec:kb}
This section focuses on the creation process of the benchmark knowledge base that addresses the primary barrier to the development of MTL research. As mentioned in previous sections, the meta-space, represented as \textit{Meta-Dataset} in Figure,\ref{fig:mtmodel_process}, needs to be established in order to construct a predictive meta-model. To accomplish this, meta-data needs to be extracted from datasets and from the executions of classification algorithms on these datasets. We obtained hundreds of datasets and extracted their characteristics, ran various algorithms on them to obtain evaluation measures, and used this information to create the meta-knowledge base. The following sections provide more details on the datasets used and their meta-features, as well as the performance evaluation and construction of the meta-knowledge base.

To simplify the process, we focus on creating the meta-dataset for a single predictive metric, such as accuracy. In Algorithm\,\ref{alg:meta_kb}, we begin by extracting the meta-features of the dataset in line\,7. Then, we apply different classification algorithms with various hyperparameter configurations and record their performance measures, such as accuracy, in line\,10. These measures are the meta-response, which when combined with the meta-features of the datasets, form the complete set of meta-data as shown in line\,11.

\begin{algorithm*}[h!]
	\caption{Establish the meta-knowledge base.}\label{alg:meta_kb}
	\begin{algorithmic}[1]
		\State  \textbf{Input}: $ClassificationAlgs[..]$,\Comment{available classification algorithms}\newline
		$HpSpace[..]$,\Comment{set of HPs configurations to be applied}\newline
		$PerfMeasures[..]$ \Comment{set of performance measures to acquire}
		\State  \textbf{Output}: meta\_KB[\#measure][\#metadata]    \Comment{meta-knowledge base}
		\Function{CreateMetaKB}{datasets[\,]}
		\State $metadata[\,] =  \varnothing  $
		\ForEach {$ \texttt{\textit{measure}}\,\,in\,\,PerfMeasures $}
		\ForEach {$ \texttt{\textit{dataset}} \,\, \texttt{DS}\,\,in\,\,datasets $}
		\State $ds\_mf =  ComputeMetaFeatures(\texttt{DS})  ;$  \Comment{See Table\,\ref{tab:metafeatures} }
		\ForEach {$ \texttt{\textit{algorithm}} \,\, \,\,\texttt{Alg}\,\,in\,\,ClassAlgs $}
		\ForEach {$ \texttt{\textit{hyperparameters\_configuration}} \,\, \texttt{Hp} \,\,in\,\,HpSpace $}
		\State $ds\_pm = GetPerformanceWith5FoldCV(\texttt{Alg},\texttt{Hp},\texttt{DS});$
		\State $metadata[\,]   \leftarrow ds\_mf \cup ds\_pm ;$
		\EndFor
		\EndFor
		\EndFor
		\State $meta\_ds[\texttt{\textit{measure}}] \leftarrow metadata[\,];$
		\EndFor
		
		\State \Return $meta\_KB$
		\EndFunction
	\end{algorithmic}
\end{algorithm*}

\subsection{Datasets}
\label{datasets}
To construct the knowledge-base, we used 400 real-world classification datasets that have been collected from the popular Kaggle\footnote{\url{https://www.kaggle.com/}}, KEEL\footnote{\url{https://sci2s.ugr.es/keel/datasets.php}}, UCI\footnote{\url{https://archive.ics.uci.edu/}}, and OpenML\,\cite{vanschorenOpenMLNetworkedScience2014} platforms. These datasets represent a mix of binary\,(71\%) and multiclass\,(29\%) classification tasks, which are highly diverse in terms of dimensionality, and class imbalance. The datasets characteristics are summarized in the Table\,\ref{tab:2}.

We utilized a total of 400 classification datasets from well-known platforms such as Kaggle, KEEL, UCI\, and OpenML to construct our knowledge base. These datasets comprise both binary,(71\%) and multiclass,(29\%) classification tasks and exhibit a wide range of characteristics, including varying levels of dimensionality and class imbalance. Table\,\ref{tab:2} provides a summary of the key features of these datasets.

\begin{table}[h!]
\caption{Statistics about the used datasets according to the number of classes, predictive attributes and instances.}
\label{tab:2}  
\centering
\begin{tabular}{p{1cm}K{1.5cm}K{2cm}K{2cm}}
	\hline
	\noalign{\smallskip}
	&Classes&	Attributes	&Instances\\
	\noalign{\smallskip}\hline\noalign{\smallskip}
	Min	&2&3&	185\\
	Max&	18&	71&	494051\\
	\noalign{\smallskip}\hline                                              
\end{tabular}
\end{table}

It might be useful to mention that the datasets utilized in this study span across various domains and application areas. To ensure fairness in the performance evaluation, no preprocessing operations were conducted on the datasets, to avoid any potential biases or impact on the classifiers' performances. The full list of datasets and their characteristics can be found on the Github repository at \url{https://github.com/LeMGarouani/AMLBID}.
 
\subsection{Meta-features}
\label{sec:used_mf}
In this work, we aim to characterize the complexity of datasets using meta-features, which are functions that extract relevant characteristics from a dataset. These meta-features produce a vector of numerical values that describe the dataset. A total of 41 meta-features were used in this study, which were extracted from the training datasets using PyMFE\,\cite{alcobacaMFEReproducibleMetafeature2020}. The meta-features were selected based on their ability to capture the dataset's complexity using measures such as Simple, Statistical, Information-theoretic, Model-based, Landmarking, and Data complexity. The list of specific meta-features used in this work is provided in Table,\ref{tab:metafeatures}.

\begin{table*}[h!]
	\caption{A sample list of meta-features used to characterize the tasks.}
	\label{tab:metafeatures}  
	%\footnotesize  % or small, footnotesize, scriptsize, tiny, etc.
\centering
		\begin{tabular}{cp{11.5cm}}
		\hline \noalign{\smallskip}
		Type&	Dataset characterization measures\,(Meta-features)\\\hline
		\noalign{\smallskip}\noalign{\smallskip}
		&\textbf{1. Simple, Statistical \& Information Theoretic}\\
		\noalign{\smallskip}\hline\noalign{\smallskip}
		Simple &	Number of instances, Number of Attributes, Number of target concept values, Proportion of minority target, Proportion of majority target, Proportion of binary attributes, Proportion of nominal attributes, Proportion of numeric attributes, Proportion of instances with missing values, Proportion of missing values, Geometric mean, Harmonic mean,\\
		Statistical & Kurtosis of data based on numerical attributes, Maximum eigenvalue Skewness of data based on numerical attributes, Covariance.\\
		Info theo& Class entropy, Uncertainty coeff.\\\hline
	
 		\noalign{\smallskip}\noalign{\smallskip}
		&\textbf{2. Model Based Measures}\\
		\noalign{\smallskip}\hline\noalign{\smallskip}
		
		Model Based & Height of tree, width of tree, Number of nodes in tree,number of leaves in tree, maximum number of nodes at one level, mean of the number of nodes on levels, length of the longest branch, Model Based length of the shortest branch, Mean of the branch lengths, Standard deviation of the branch lengths, Minimum occurrence of attributes, Maximum occurrence of attributes, Mean of the number of occurrences of attributes, Standard deviation of the number of occurrences of attributes 
		\\\hline

 	\noalign{\smallskip}\noalign{\smallskip}
	&\textbf{3. Landmarking Based Measures}\\
	\noalign{\smallskip}\hline\noalign{\smallskip}
 
	Landmarking & Naive Bayes, ii) 1-NN (Nearest Neighbor), iii) Elite 1-NN, iv) decision Tree learner and v) a random chosen node learner.	\\\hline
	
	\noalign{\smallskip}\noalign{\smallskip}
	&\textbf{ 4. Complexity Based Measures}\\
	\noalign{\smallskip}\hline\noalign{\smallskip}
	Dimentionality &Average number of points per dimension, Ratio of the PCA dimension to the original dimension\\
	Class balance &	Entropy of classes proportions, Imbalance ratio\\
		
		\noalign{\smallskip}\hline                                              
	\end{tabular}
\end{table*}

\subsection{Meta-dataset}
 \label{mkb_construction}
In order to construct the meta-knowledge base, we employed eight classifiers from Scikit-learn\footnote{\url{https://scikit-learn.org}}. The classifiers used in this study are Support Vector Machines, Logistic Regression, Decision Tree, Random Forest, Extra Trees, Gradient Boosting, AdaBoost, and Stochastic Gradient Descent. A comprehensive description of the algorithms and their hyperparameters can be found in Table\,\ref{tab:9}.

\begin{table*}[h!]
	\caption{Hyperparameters tuned in the experiments. }
	\label{tab:9}       % Give a unique label
	\begin{tabular}{llp{0.23\linewidth}p{0.55\linewidth}}
		\hline\noalign{\smallskip}
		Algo.&Hyperparameter & Values & Description  \\\hline\noalign{\smallskip\smallskip}

\parbox[t]{2mm}{\multirow{5}{*}{\rotatebox[origin=c]{90}{LR}}}&	C & 	[$1e^{-10}$, 10.] (log-scale) & 	Regularization strength.\\
	&	penalty  & 	\{’l2’, ’l1’ \} & 	Whether to use Lasso or Ridge regularization.\\
	&	Fit\_intercept &  	{True, False} & 	Whether or not the intercept of the linear classifier should be computed.\\\hline\noalign{\smallskip}
	
	\parbox[t]{2mm}{\multirow{5}{*}{\rotatebox[origin=c]{90}{DT}}}&	max features & 	[0.1, 0.9]	 & Number of features to consider when computing the best node.\\
	&		min\_samples\_leaf  & 	[1, 21]	 & The min nbr of samples required to be at a leaf node.\\
	&	Min\_samples\_split & 	[2, 21]	 & The min nbr of samples required to split an internal node.\\
	&	criterion & 	\{’entropy’, ’gini’ \} & 	Function used to measure the quality of a split.\\\hline\noalign{\smallskip}

	\parbox[t]{2mm}{\multirow{5}{*}{\rotatebox[origin=c]{90}{SVM\;\;}}}&complexity  & [$1e^{-10}$, 500] (log-scale) & Soft-margin constant, controlling the trade-off between model simplicity and model fit. \\
	& Kernel & \{'poly', 'rbf'\}  & The function of kernel is to take data as input and transform it into the required form (linear, nonlinear, polynomial, etc.). \\
	& 	coef0 &[0., 10] & Additional coefficient used by the sigmoid kernel. \\
	&	gamma &  [$1e^{-3}$, 1.01] (log-scale)& Length-scale of the kernel function, determining its locality. \\
	&	Degree  & [2, 3] & Degree for the `poly' kernel. \\\hline\noalign{\smallskip}
	
	\parbox[t]{2mm}{\multirow{5}{*}{\rotatebox[origin=c]{90}{SGD Classifier\;\;\;\;\;}}}&	loss & 	\{’hinge’,’perceptron’,’log’, ’squared\_hinge’\} & 	Loss function to be optimized.\\
	&	penalty	 & \{’l2’, ’l1’, ’elasticnet’ \} & 	Whether to use Lasso, Ridge, or ElasticNet regularization.\\
	&	learning rate & 	\{’const’, ’opt’, ’invscaling’ \} & 	Shrinks the contribution of each successive training update.\\
	&	fit intercept  & 	\{True, False\} & 	Whether or not the intercept of the linear classifier should be computed.\\
	&	l1 ratio & 	[0., 1.] & 	Ratio of Lasso vs. Ridge regularization to use.
	Only used when the `penalty' is ElasticNet.\\
	&eta0 & 	[0., 5.] & 	Initial learning rate.\\\hline
	
	\parbox[t]{2mm}{\multirow{5}{*}{\rotatebox[origin=c]{90}{RF \& ET\;\;}}}&	bootstrap & 	\{true, false\}	 & Whether to train on bootstrap samples or on the full set. \\
	&		Max\_features & 	[0.1, 0.9] & 	Fraction of random features sampled per node. \\
	&		Min\_samples\_leaf & 	[1, 20]	 & The min number of data points required in order to create a leaf. \\
	&	Min\_samples\_split & 	[2, 20]	 & The min number of data points required to split an internal node.\\
	&	imputation & 	{mean, median, mode} &  	Strategy for imputing missing numeric variables.\\
	&	split criterion	 & \{entropy, gini\} & 	Function to determine the quality of a possible split.\\\hline\noalign{\smallskip}
	
	\parbox[t]{2mm}{\multirow{5}{*}{\rotatebox[origin=c]{90}{Grad. Boosting}}}&		Learning\_rate & 	[0.01, 1] & 	Shrinks the contribution of each successive DT in the ensemble.  \\
	&	criterion & 	\{’friedman\_mse’, ’mse’ \} & 	The function to measure the quality of a split.  \  \\
	&	N\_estimators & 	[50, 501] & 	Number of decision trees in the ensemble.  \\
	&	max depth & 	[1, 11] & 	Controls the complexity of the decision trees. \\
	&Min\_samples\_split & 	[2, 21] & 	The minimum number of samples required to split an internal node. \\	\noalign{\smallskip}\hline
	
	 \parbox[t]{2mm}{\multirow{5}{*}{\rotatebox[origin=c]{90}{\;Adaboost}}}&	algorithm & 	\{SAMME, SAMME.R\} & 	Determines which boosting algorithm to use.\\
&		N\_estimators & 	[50, 501] & 	Number of estimators to build.\\
&		learning rate & 	[0.01, 2.0] (log-scale) & 	Learning rate shrinks the contribution of each classifier.\\
	&	Max\_depth & 	[1, 11] & 	The maximal depth of the decision trees.\\\hline\noalign{\smallskip}

	\end{tabular}
\end{table*}

The knowledge base represents the collective experience gained from previous optimizations of the classifiers on the 400 datasets. It is comprised of the meta-features $m_1,...,m_{400}$ and the corresponding optimal hyperparameter values of the classifiers for each dataset. In other words, it can be expressed as\,: $\mathcal{K}_B= \{(m_1, A^{(1)}_{H^1}),\dots,(m_{400}, A^{(n)}_{H^n})\}$.

To gain a better understanding of how the knowledge base is constructed, let us consider the execution scenario of the proposed framework. For each execution of an algorithm $\textbf{A}$ on each dataset $\textbf{D}$, we generate at least 1000 different combinations of hyperparameter configurations. This execution process results in an average of 8000 pipelines for each dataset. It's worth noting that to ensure stable performance, we performed a $10\times5$-fold stratified cross-validation strategy for the estimation of pipeline performance during the construction of the meta-datasets. This process controls the variation caused by different choices of training and test instances. As a result, the knowledge base consists of over 4 million evaluated classification pipelines, considering the performance measures of Accuracy, Recall, F1-score, and Precision. The number of configurations/evaluations of each considered algorithm is not the same due to the different variations of algorithm hyperparameters. Finally, we obtained a meta-dataset for each classification algorithm and performance measure, which is fed into the Meta-knowledge base.

\subsubsection{The meta-knowledge base schema}
\begin{figure*}[h!]
	\centering
	\includegraphics[width=1\textwidth]{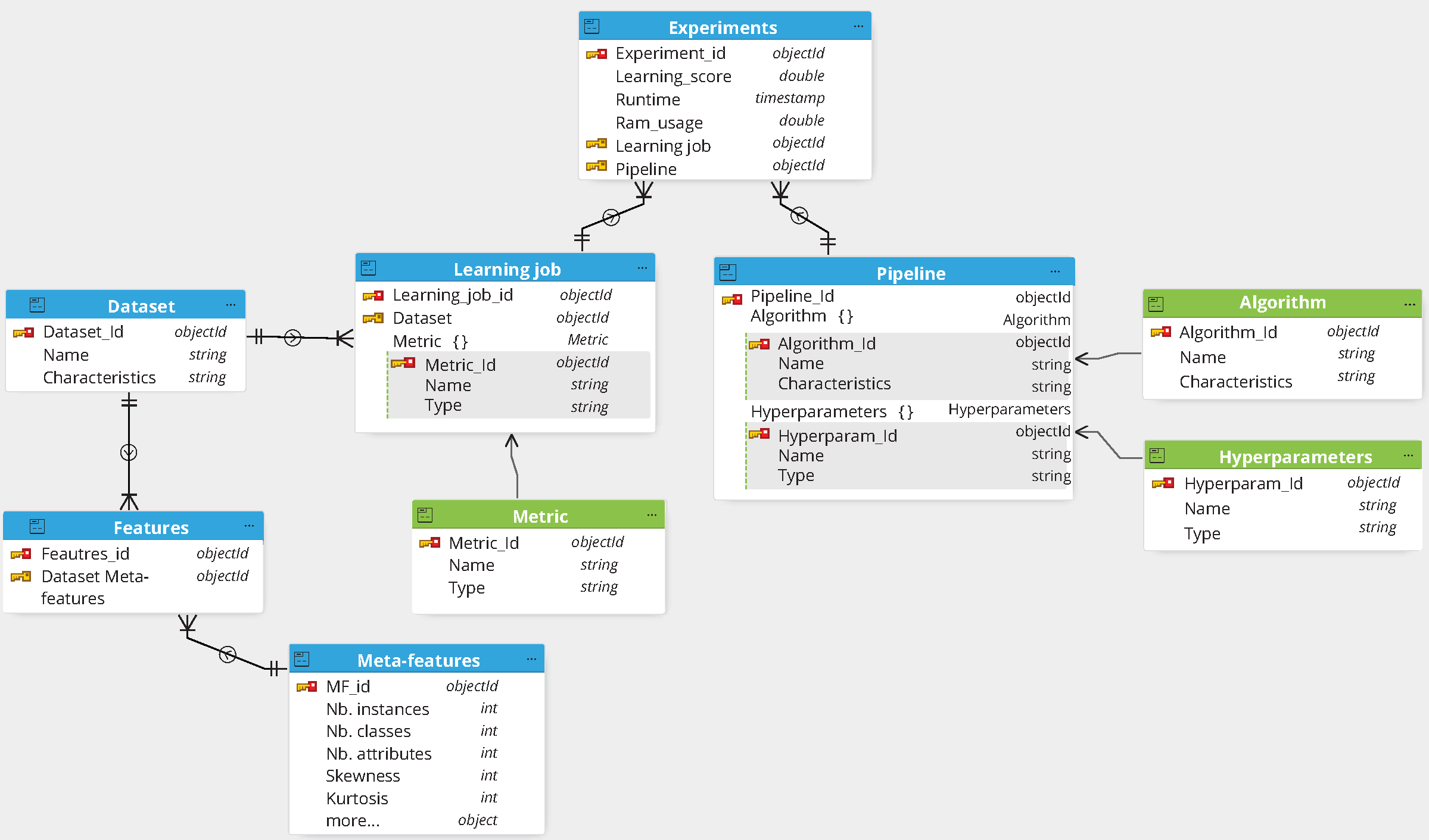}
	\caption{The schema of the knowledge base.}
	\label{fig:KBSchema}       % Give a unique label
\end{figure*}

The schema of the meta-knowledge base is presented in Figure\,\ref{fig:KBSchema}. The knowledge base is designed to store the outcomes of the experiments and enable easy access to any operation and data during the inference phase. As the KB is primarily utilized for storing and extracting data without the need for complex queries, and since each dataset, pipeline, and experiment may differ from one another, it can be structured in a NoSQL database format. This is similar to what was implemented in\,\cite{garouthesis2022}, where the schema includes the entities that serve as the foundation for our KB.

The knowledge base consists of several entities which are used to store and manage data during the inferring phase. The entities, based on a NoSQL database format used in previous works,\cite{garouthesis2022}, are:
\begin{itemize}
	\item \textbf{Datasets.} Each dataset is associated with its corresponding meta-features metadata and a learning job.
    \item \textbf{Learning jobs.} A learning job is a combination of a dataset and a metric, and serves as input for a given ML pipeline.
    \item \textbf{Pipelines.} The pipelines are pre-generated and stored to be used in experiments later on.
    \item \textbf{Hyperparameters.} Hyperparameter values are determined when associated with an algorithm through the Pipeline entity.
    \item \textbf{Experiments.} Results from experiments include pipeline scores on the defined learning job's metrics, runtime, and RAM usage.
\end{itemize}

%% refer to the ERD schema, characteristics is an elementary column or list ?
% characteristics in dataset is int and algorithm is string column where is in pipline is sub-component (sub-class) that doesn't exist in relational erd but in OO design. the r/s in ERD are always bi-directional and associations doesn't give hierarchical structures

%The collection of experiments can be used for pre-training our model’s network on predefined pipelines.

\section{Empirical evaluation}
\label{sec:experiments}
The performance of the algorithms and a range of dataset characteristics are stored to build the meta-space and the meta-knowledge base. In order to do this, it is first converted into a learning dataset. Then, a meta-model is trained on it to capture the connections between dataset characteristics and performance data of potential algorithms. The mapping of the meta-model is designed to learn the intricate relationships between meta-features of a task and the effectiveness of the corresponding ML pipelines. The aim is to suggest the most appropriate configuration of ML algorithm(s) based on the meta-features of a new task.

In this context, a task\,(i.e., a dataset) $t_j$ from the set of known tasks $\mathcal{T}$ can be represented as a vector $F(t_j) = (m_{j,1}, \ldots, m_{j,K})$ of $K$ meta-features $m_{j,K} \in F$. This vector can be used to calculate the similarity between $t_{new}$ (i.e., a new task) and $t_j$ using a distance metric such as the Euclidean distance. This similarity calculation allows for information transfer from the most similar tasks to the new task $t_{new}$. The Euclidean distance between the meta-features of $t_{new}$ and $t_j$ can be computed using Equation\,(\ref{eq:1}):

\begin{equation}\label{eq:1}
d\left( m(t_{new}),m(t_j)\right) = \sqrt {\sum {i=1}^{k} \left( m{t_{new, i}}-m_{t_{j, i}}\right)^2 }
\end{equation}

The goal is to create an enhanced meta-model that can suggest the best performing classification configuration(s) for an unseen dataset and a classification evaluation measure. To achieve this objective, a few fundamental criteria are considered when selecting the appropriate meta-learner. First, the problem in the meta-learning space is a classification problem where a class needs to be predicted, which is whether a given ML pipeline is among the best performing pipelines or not. The second criterion is that the meta-model must be more sensitive, meaning that it should be able to identify even the subtle meta-features that characterize the datasets. This is because we need to predict how the hyperparameters' values affect the data mining results and establish the correlation between the datasets' meta-features and the different ML algorithm types and configurations. The third criterion is that the meta-learner should be able to handle missing values. It is worth noting that some dataset characteristics\,(meta-features) can only be calculated on datasets that contain either continuous or categorical attributes. Hence, we select two state-of-the-art learning algorithms, Random Forest\,(RF) and k-Nearest Neighbor\,(kNN), to generate meta-models that can predict the most appropriate pipelines for a given dataset.

The k-Nearest Neighbor algorithm, known for its effectiveness and often referred to as "nearest neighbor imputation," as well as the Random Forest learner that meets all the above-mentioned criteria, have been chosen. RF suffers far less from the discreteness of the leaves because many trees\,(i.e., 500 trees) are built at random, and averages are taken to be used as predictions. It also performs well when missing values are present. Therefore, we use kNN and RF to construct meta-models for each data mining performance measure that we consider.

%Thus, when the meta-learning system is applied to a new dataset, the meta-model returns a ranking of the most suitable classification algorithms and configurations, based on its meta-feature values.

To obtain a top-K ranking, the KNN classifier is often used for ranking. When a new dataset is introduced to the meta-learning system, the kNN identifies the k-nearest datasets\,(KND) to the candidate dataset in the meta-knowledge base by employing the Euclidean distance measure\,(Eq.\,(\ref{eq:1})). Using this measure, a vector $d = [d_{1},d_{2}, \dots,d_{k}]$ containing the dissimilarities among all characteristics\,(meta-features) of the datasets is constructed. A weighted average of each individual neighbor is then used to predict the optimal pipeline configuration based on the relevant measure. The k-nearest datasets selection approach is presented as pseudo code in Algorithm\,\ref{alg:KND}.

\begin{algorithm} 
	\caption{K-nearest datasets selection.}\label{alg:KND}
	\begin{algorithmic}[1]
		\State  \textbf{Input}:  Ds[..],\Comment{new dataset chosen by the user}\newline
		meta\_KB[..]  \Comment{the constructed knowledge base}
		%KnowledgeBase[..],\Comment{the constructed Knowledge base}\newline
		
		\State  \textbf{Output}: KND[..]  \Comment{K-nearest datasets distance vector} 
		
		\Function{KND\_Selection}{Ds[\,]}
		\State $KND =  \varnothing  $
		\State $ds\_mf =  ComputeMetaFeatures(\texttt{Ds}) $
		\ForEach {$ \texttt{dataset d \textit{in} meta\_KB} $}
		\State $d \leftarrow  \sqrt {\sum _{i=1}^{k}  \left( m_{DS}-m_{\textit{d}}\right)^2}$
		\State $KND \leftarrow d $
		\EndFor		
		\State \Return $\texttt{K-nearest datasets distance vector}$
		\EndFunction
		
	\end{algorithmic}
\end{algorithm}

\subsection{The experimental configuration}
In order to ensure a fair and informative comparison, we evaluate our approach on a diverse set of 30 datasets, which include both binary and multiclass classification problems. These datasets are obtained mainly from the OpenML AutoML benchmark\,\cite{gijsbersOpenSourceAutoML2019}, and they represent a wide range of classification problems with varying sample sizes and dimensionalities. It is worth noting that these 30 datasets have not been used by any learning algorithm during the offline phase of our meta-learning system.

\begin{table}[h!]
	\caption{List\,(sample) of datasets used in the evaluation.}
	\label{tab:c5}  
	\centering
	\begin{tabular}{p{1.5cm}K{2.3cm}K{2.5cm}}
		\hline
		\noalign{\smallskip}
		Dataset	&  Num Class.&	Num Inst.\\
		\noalign{\smallskip}\hline\noalign{\smallskip}
		\cite{mazumderFailureRiskAnalysis2021b}&	4&	959\\
		\cite{saravanamuruganChatterPredictionBoring2017}&	3&	2000\\
		\cite{benkedjouhHealthAssessmentLife2015}&	2&	61000\\

		APSFailure&	2&	60000\\
		Higgs&	2&	110000	\\
		CustSat	&2&	76020\\
		
		\noalign{\smallskip}\hline                                              
	\end{tabular}
\end{table}
%%%%%%%%%%%%%
%\subsubsection{The evaluation method}
We trained the recommended ML pipelines on the benchmark datasets and compared their performance with that of the TPOT and Auto-sklearn frameworks. For TPOT, we used the default settings, i.e., generating and evaluating 100 pipelines for each dataset. In contrast, for Auto-sklearn, we compared Meta-learning with two versions, namely Auto-sklearn(V) that produces a single optimal pipeline and Auto-sklearn(E) that generates a set of 50 best pipelines.

To ensure fairness in performance comparison, we executed all AutoML systems on our local hardware, which consists of an Intel(R) Core(TM) i9-10900KF CPU @ 3.70GHz and 32 GB RAM. We utilized the predefined settings to divide each dataset into 5 stratified folds and run each tool on 10 CPU cores to produce the final pipeline.

%It is important to note that the both baseline frameworks evaluate each generated pipeline by running that on the given dataset, whereas AMLBID immediately produces, at an imperceptible computational cost, a list of potential top pipelines configurations using its meta-knowledge base.

\subsection{Experimental results}

In our paper\,\cite{garouaniUsingMetalearningAutomated2022}, we provide a comparative evaluation of the results obtained from the 30 benchmark datasets using the MtL framework, TPOT, and two versions of Auto-sklearn. The results indicate that the MtL-based AutoML outperforms the baseline models, even though no pipelines were executed prior to the recommendation. The evaluation also highlights the superior accuracy of the meta-model with a rich knowledge base compared to TPOT and Auto-sklearn.
%. We count how many times the system performs better ($>$) or worse ($<$)
\begin{table}[h!]
	\caption[Performance of AutoML systems on the 30-benchmark datasets, comparing the proposed system with TPOT and two versions of Auto-sklearn.]{Performance of AutoML systems on the 30-benchmark datasets. The table includes the number of times each system performs better\,($>$) or worse\,($<$) than the others.}
	\label{tab:c56}
	\centering  
	\begin{tabular}{p{3.5cm}K{1.5cm}K{2cm}}
		\hline
		\noalign{\smallskip}
		System	&$>$	&$<$ \\
		\noalign{\smallskip}\hline\noalign{\smallskip}
		Meta-learning&	\textbf{19}&	\textbf{2}\\
		TPOT&	6&	5\\
		Auto-sklearn&	5&	23\\

		\noalign{}\hline                                              
	\end{tabular}
\end{table}

Most state-of-the-art AutoML systems suffer from the drawback of high computational complexity, often requiring a significant amount of time and resources to perform on non-conventional datasets. In contrast, MtL has the advantage of having an O(1)\,\cite{garouaniUsingMetalearningAutomated2022} computational complexity, allowing for generating recommendations in a negligible amount of time.% This argument is further supported by the results presented in Table\,\ref{tab:8}, which compares the execution time of Meta-learning, TPOT, and Auto-sklearn on the same machine for the benchmarked datasets.

The significant performance advantage of MtL over other AutoML approaches can be attributed to the fact that other approaches require extensive training time to evaluate multiple algorithms with various configurations on the same dataset in order to produce recommendations. On the other hand, as explained in section\,\ref{mkb_construction}, the MtL system utilizes a pre-existing meta-knowledge base with over 4 million evaluated pipelines, allowing it to generate recommendations through comparative search of meta-features with the most similar existing datasets. Moreover, with each iteration of new datasets, the meta-knowledge base can be further enriched through evolutionary training.

We opted to use the k-nearest neighbor meta-model for comparison with the decision tree meta-model, as the kNN classifier is one of the most commonly employed algorithms for obtaining top-k rankings in meta-learning\,\cite{garouaniAutomationIndustrialData2021b}. We used the Euclidean Distance metric and forecasted the optimal pipeline configuration by taking a weighted average of each individual neighbor's ranking in order to determine the closest neighbors of the dataset. %The choice of kNN meta-model over the random forest can be supported by the results presented in Figure\,\ref{fig:3}, which illustrates the performance of the RF and kNN meta-models in recommending the optimal predictive pipeline configuration. 
According to the accuracy metric, the kNN-based meta-model outperforms the random forest classifier-based meta-learner.

% \begin{figure}[h!]
% 	\centering
% 	\resizebox{1\hsize}{!}{\includegraphics{Figures/Fig3.png}}
% 	% figure caption is below the figure
% 	\caption{Predictive performance of the kNN and RF meta-models.}
% 	\label{fig:3}       % Give a unique label
% \end{figure}

\section{Challenges and open directions}
\label{sec:future_directions}

Meta-learning systems utilize a collection of data attributes to describe and characterize data mining tasks, aiming to uncover the relationships between attributes and the performance of learning algorithms. The accurate identification of these data properties is crucial for effectively mapping tasks onto learning mechanisms. As a data-centric approach, the efficacy of meta-learning is significantly influenced by the quality of task descriptions, or meta-features, which must adequately capture the characteristics of the primary learning tasks or datasets in order to enable effective knowledge transfer across them.

Meta-learning effectiveness heavily relies on the quality of the task description, represented by meta-features. In this regard, several approaches in meta-learning utilize families of meta-features to measure task similarity, commonly quantified using Euclidean distance. However, such methods have shown limitations, particularly in identifying relevant meta-features\,\cite{vanschorenMetaLearningSurvey2018}. This raises important research questions such as "What criteria should be used to select or exclude a family of meta-features?" For instance, statistical meta-features may lack expressiveness and not always be intuitive. In addition, different datasets can share identical statistical properties but have different data distributions\,\cite{matejkaSameStatsDifferent2017}. The selection of meta-features is usually an \textit{ad hoc} process based on domain knowledge. Therefore, developing more predictive and informative meta-features is crucial for enhancing meta-learning effectiveness.

In our view, conventional meta-features may not always capture the critical attributes of a specific task, despite some of them being highly task-specific\,\cite{meskhiLearningAbstractTask2021}. This is because they model only the general characteristics of the dataset, such as the number of instances and features, class imbalance, and so on. Therefore, developing relevant meta-features could prove useful in identifying the latent relationships across tasks, and building accurate meta-models\,\cite{amlbid}.

\section{Conclusion}

The selection and parameterization of machine learning algorithms is a complex and time-consuming task. Motivated by the academic dream and industrial needs, the automated machine learning has recently became a hot topic in order to relieve the analysts\,(either experts or novices) from the ML pipeline building difficulties. This paper presents a comprehensive and experimental review of current AutoML approaches and tools, focusing on Meta-learning as a promising paradigm for addressing the problem of algorithms selection and configuration. In this work, we first define the problem in contexts of AutoML, and then introduce the fundamental concept of AutoML and the related tools and techniques to cope the problem. Additionally, the paper discusses taxonomies of existing works based on "what" and "how" to automate, which serve as valuable guidelines for designing new approaches and utilizing existing AutoML techniques. The study also categorizes related research on MtL into three general groups based on the ultimate objectives of MtL systems. We further review state-of-art AutoML tools and platforms. Finally, we survey the potential of these tools on industrial big data.

Despite the increasing number of related works in recent years, there are still various aspects that require additional investigation. Existing literature demonstrates certain trends\,: most studies have generated meta-knowledge by executing Grouping Strategies with Holdout or single Cross-Validation resampling techniques, characterizing datasets using simple and statistical meta-features, and recommending Hyperparameters using a kNN meta-model. The reproducibility of experiments and the sharing of results remain unexplored areas. Furthermore, no comprehensive solution for the CASH problem, utilizing an end-to-end MtL resolver, has been proposed or investigated. Addressing these aspects would provide the research community with valuable meta-knowledge to inform future studies. Therefore, this research explores different experimental setups, meta-features, and procedures applied to learning tasks, potentially opening up new horizons in the field of MtL research.


 \bibliographystyle{IEEEtran}
\bibliography{Paper}

\begin{thebibliography}{100}
\providecommand{\url}[1]{#1}
\csname url@samestyle\endcsname
\providecommand{\newblock}{\relax}
\providecommand{\bibinfo}[2]{#2}
\providecommand{\BIBentrySTDinterwordspacing}{\spaceskip=0pt\relax}
\providecommand{\BIBentryALTinterwordstretchfactor}{4}
\providecommand{\BIBentryALTinterwordspacing}{\spaceskip=\fontdimen2\font plus
\BIBentryALTinterwordstretchfactor\fontdimen3\font minus
  \fontdimen4\font\relax}
\providecommand{\BIBforeignlanguage}[2]{{%
\expandafter\ifx\csname l@#1\endcsname\relax
\typeout{** WARNING: IEEEtran.bst: No hyphenation pattern has been}%
\typeout{** loaded for the language `#1'. Using the pattern for}%
\typeout{** the default language instead.}%
\else
\language=\csname l@#1\endcsname
\fi
#2}}
\providecommand{\BIBdecl}{\relax}
\BIBdecl

\bibitem{Garouani_sncs}
M.~Choaib, M.~Garouani, M.~Bouneffa, and al., ``Iot-aid: An automated decision
  support framework for iot,'' \emph{SN Computer Science}, vol.~5, no.~4, April
  2024.

\bibitem{caldarelliSignalBasedApproachNews2016}
S.~Caldarelli, D.~F. Gurini, A.~Micarelli, and G.~Sansonetti, ``A
  {{Signal-Based Approach}} to {{News Recommendation}},'' in \emph{{{UMAP}}},
  2016.

\bibitem{biancalanaContextawareMovieRecommendation2011}
S.~Jayalakshmi, N.~Ganesh, R.~{\v{C}}ep, and J.~S. Murugan, ``Movie recommender
  systems: Concepts, methods, challenges, and future directions,''
  \emph{Sensors}, vol.~22, no.~13, p. 4904, jun 2022.

\bibitem{onoriComparativeAnalysisPersonalityBased2016}
M.~Onori, A.~Micarelli, and G.~Sansonetti, ``A {{Comparative Analysis}} of
  {{Personality-Based Music Recommender Systems}},'' in
  \emph{{{EMPIRE}}@{{RecSys}}}, 2016.

\bibitem{sansonettiEnhancingCulturalRecommendations2019}
G.~Sansonetti, F.~Gasparetti, A.~Micarelli, F.~Cena, and C.~Gena, ``Enhancing
  cultural recommendations through social and linked open data,'' \emph{User
  Modeling and User-Adapted Interaction}, vol.~29, no.~1, pp. 121--159, 2019.

\bibitem{sansonettiPointInterestRecommendation2019}
G.~Sansonetti, ``Point of interest recommendation based on social and linked
  open data,'' \emph{Personal and Ubiquitous Computing}, vol.~23, no.~2, pp.
  199--214, 2019-04-01.

\bibitem{fogliExploitingSemanticsContextaware2019}
A.~Fogli and G.~Sansonetti, ``Exploiting semantics for context-aware itinerary
  recommendation,'' \emph{Personal and Ubiquitous Computing}, vol.~23, no.~2,
  pp. 215--231, 2019-04-01.

\bibitem{kulkarniTrafficLightDetection2018}
R.~Kulkarni, S.~Dhavalikar, and S.~Bangar, ``Traffic {{Light Detection}} and
  {{Recognition}} for {{Self Driving Cars Using Deep Learning}},'' 2018, pp.
  1--4.

\bibitem{CoDIT_choaib}
M.~Choaib, M.~Garouani, M.~Bouneffa, and Y.~Mohanna, ``Iot sensor selection in
  cyber-physical systems: Leveraging large language models as recommender
  systems,'' in \emph{2024 10th International Conference on Control, Decision
  and Information Technologies (CoDIT)}, 2024, pp. 2516--2519.

\bibitem{edas}
M.~Chaabi, M.~Hamlich, and M.~Garouani, ``Product defect detection based on
  convolutional autoencoder and one-class classification,'' \emph{{IAES}
  International Journal of Artificial Intelligence}, vol.~12, pp. 912--920,
  Oct. 2022.

\bibitem{Garouani2022a}
M.~Garouani, A.~Ahmad, M.~Bouneffa, M.~Hamlich, G.~Bourguin, and
  A.~Lewandowski, ``Towards big industrial data mining through explainable
  automated machine learning,'' \emph{The International Journal of Advanced
  Manufacturing Technology}, vol. 120, no. 1-2, pp. 1169--1188, Feb. 2022.

\bibitem{10849387}
M.~Garouani, J.~Mothe, A.~Barhrhouj, and J.~Aligon, ``Investigating the duality
  of interpretability and explainability in machine learning,'' in \emph{2024
  IEEE 36th International Conference on Tools with Artificial Intelligence
  (ICTAI)}, 2024, pp. 861--867.

\bibitem{khan_2020}
I.~Khan, X.~Zhang, M.~Rehman, and R.~Ali, ``A literature survey and empirical
  study of meta-learning for classifier selection,'' \emph{IEEE Access},
  vol.~8, p. 10262–10281, 2020.

\bibitem{garouaniAutomationIndustrialData2021}
M.~Garouani, A.~Ahmad, M.~Bouneffa, A.~Lewandowski, G.~Bourguin, and
  M.~Hamlich, ``Towards the {{Automation}} of {{Industrial Data Science}}: {{A
  Meta-learning}} based {{Approach}},'' 2021-08-10, pp. 709--716.

\bibitem{koberReinforcementLearningRobotics2013}
J.~Kober, J.~A. Bagnell, and J.~Peters, ``Reinforcement learning in robotics:
  {{A}} survey,'' \emph{The International Journal of Robotics Research},
  vol.~32, no.~11, pp. 1238--1274, 2013.

\bibitem{greenspanGuestEditorialDeep2016}
H.~Greenspan, B.~van Ginneken, and R.~M. Summers, ``Guest editorial deep
  learning in medical imaging: Overview and future promise of an exciting new
  technique,'' \emph{{IEEE} Transactions on Medical Imaging}, vol.~35, no.~5,
  pp. 1153--1159, may 2016.

\bibitem{chenDeepDrivingLearningAffordance2015}
C.~Chen, A.~Seff, A.~Kornhauser, and J.~Xiao, ``{{DeepDriving}}: {{Learning
  Affordance}} for {{Direct Perception}} in {{Autonomous Driving}},'' in
  \emph{2015 {{IEEE International Conference}} on {{Computer Vision}}
  ({{ICCV}})}, 2015-12, pp. 2722--2730.

\bibitem{10.1007/978-3-031-48232-8_42}
M.~Garouani and M.~Bouneffa., ``Unlocking the black box: Towards interactive
  explainable automated machine learning,'' in \emph{Intelligent Data
  Engineering and Automated Learning. IDEAL 2023}.\hskip 1em plus 0.5em minus
  0.4em\relax Cham: Springer Nature Switzerland, 2023, pp. 458--469.

\bibitem{Garouani_WISE2024}
M.~Garouani, F.~Ravat, and N.~Valles-Parlangeau, ``Model lake : A new
  alternative for machine learning models management and governance,'' in
  \emph{Web Information Systems Engineering -- WISE 2024}, M.~Barhamgi,
  H.~Wang, and X.~Wang, Eds.\hskip 1em plus 0.5em minus 0.4em\relax Singapore:
  Springer Nature Singapore, 2025, pp. 133--144.

\bibitem{pria}
M.~Garouani and M.~Bouneffa, ``Automated machine learning hyperparameters
  tuning through meta-guided bayesian optimization,'' \emph{Progress in
  Artificial Intelligence}, Jan. 2024.

\bibitem{cohen-shapiraAutoGRDModelRecommendation2019}
N.~Cohen-Shapira, L.~Rokach, B.~Shapira, G.~Katz, and R.~Vainshtein,
  ``{{AutoGRD}}: {{Model Recommendation Through Graphical Dataset
  Representation}},'' in \emph{Proceedings of the 28th {{ACM International
  Conference}} on {{Information}} and {{Knowledge Management}}}, ser. {{CIKM}}
  '19, 2019-11-03, pp. 821--830.

\bibitem{feurerEfficientRobustAutomated2015a}
M.~Feurer, A.~Klein, K.~Eggensperger, J.~T. Springenberg, M.~Blum, and
  F.~Hutter, ``Efficient and robust automated machine learning,'' in
  \emph{Proceedings of the 28th {{International Conference}} on {{Neural
  Information Processing Systems}} - {{Volume}} 2}, ser. {{NIPS}}'15, Dec.
  2015, pp. 2755--2763.

\bibitem{bergstraRandomSearchHyperparameter2012}
J.~Bergstra and Y.~Bengio, ``Random search for hyper-parameter optimization,''
  \emph{The Journal of Machine Learning Research}, vol.~13, pp. 281--305, 2012.

\bibitem{garouthesis2022}
\BIBentryALTinterwordspacing
M.~Garouani, ``Towards efficient and explainable automated machine learning
  pipelines design : Application to industry 4.0 data,'' Ph.D. dissertation,
  2022. [Online]. Available: \url{http://www.theses.fr/2022DUNK0620}
\BIBentrySTDinterwordspacing

\bibitem{feurerInitializingBayesianHyperparameter2015}
M.~Feurer, J.~T. Springenberg, and F.~Hutter, ``Initializing bayesian
  hyperparameter optimization via meta-learning,'' in \emph{Proceedings of the
  {{Twenty-Ninth AAAI Conference}} on {{Artificial Intelligence}}}, ser.
  {{AAAI}}'15.\hskip 1em plus 0.5em minus 0.4em\relax {AAAI Press}, 2015-01-25,
  pp. 1128--1135.

\bibitem{olsonEvaluationTreebasedPipeline2016}
R.~S. Olson, N.~Bartley, R.~J. Urbanowicz, and J.~H. Moore, ``Evaluation of a
  {{Tree-based Pipeline Optimization Tool}} for {{Automating Data Science}},''
  in \emph{Proceedings of the {{Genetic}} and {{Evolutionary Computation
  Conference}} 2016}, ser. {{GECCO}} '16, 2016-07-20, pp. 485--492.

\bibitem{andradottirReviewRandomSearch2015}
S.~Andrad{\'{o}}ttir, ``A review of random search methods,'' in
  \emph{International Series in Operations Research \&
  Management Science}.\hskip 1em plus 0.5em minus 0.4em\relax Springer New
  York, sep 2014, pp. 277--292.

\bibitem{mantovaniRethinkingDefaultValues2021}
R.~G. Mantovani, A.~L.~D. Rossi, E.~Alcoba{\c{c}}a, and J.~C.~G. and,
  ``Rethinking default values: a low cost and efficient strategy to define
  hyperparameters,'' \emph{CoRR}, vol. abs/2008.00025, 2020.

\bibitem{gomesCombiningMetalearningSearch2010}
T.~A.~F. Gomes, R.~B. C., C.~Soares, A.~L.~D. Rossi, and A.~Carvalho,
  ``Combining {{Meta-learning}} and {{Search Techniques}} to {{SVM Parameter
  Selection}},'' in \emph{2010 {{Eleventh Brazilian Symposium}} on {{Neural
  Networks}}}, 2010-10, pp. 79--84.

\bibitem{padiernaHyperParameterTuningSupport2017}
L.~C. Padierna, M.~Carpio, and A.~Rojas, ``Hyper-{{Parameter Tuning}} for
  {{Support Vector Machines}} by {{Estimation}} of {{Distribution
  Algorithms}},'' in \emph{Nature-{{Inspired Design}} of {{Hybrid Intelligent
  Systems}}}, ser. Studies in {{Computational Intelligence}}, 2017, pp.
  787--800.

\bibitem{snoekPracticalBayesianOptimization2012}
J.~Snoek, H.~Larochelle, and R.~P. Adams, ``Practical {{Bayesian Optimization}}
  of {{Machine Learning Algorithms}},'' 2012.

\bibitem{Sadasc}
M.~Garouani, A.~Ahmad, M.~Bouneffa, and M.~Hamlich, ``Scalable meta-bayesian
  based hyperparameters optimization for machine learning,'' in \emph{Smart
  Applications and Data Analysis}.\hskip 1em plus 0.5em minus 0.4em\relax
  Springer International Publishing, 2022, pp. 173--186.

\bibitem{kandasamyMultifidelityGaussianProcess2019}
K.~Kandasamy, G.~Dasarathy, J.~B. Oliva, J.~Schneider, and B.~Poczos,
  ``Multi-fidelity {{Gaussian Process Bandit Optimisation}},'' 2019.

\bibitem{maclaurinGradientbasedHyperparameterOptimization2015}
D.~Maclaurin, D.~Duvenaud, and R.~P. Adams, ``Gradient-based {{Hyperparameter
  Optimization}} through {{Reversible Learning}},'' 2015.

\bibitem{hutterSequentialModelBasedOptimization2011}
F.~Hutter, H.~H. Hoos, and K.~Leyton-Brown, ``Sequential {{Model-Based
  Optimization}} for {{General Algorithm Configuration}},'' in \emph{Learning
  and {{Intelligent Optimization}}}, ser. Lecture {{Notes}} in {{Computer
  Science}}, C.~A.~C. Coello, Ed.\hskip 1em plus 0.5em minus 0.4em\relax
  {Springer}, 2011, pp. 507--523.

\bibitem{bergstraAlgorithmsHyperparameterOptimization2011}
J.~Bergstra, R.~Bardenet, Y.~Bengio, and B.~K\'egl, ``Algorithms for
  hyper-parameter optimization,'' in \emph{Proceedings of the 24th
  {{International Conference}} on {{Neural Information Processing Systems}}},
  ser. {{NIPS}}'11.\hskip 1em plus 0.5em minus 0.4em\relax {Curran Associates
  Inc.}, 2011, pp. 2546--2554.

\bibitem{hutterAutomatedMachineLearning2019}
F.~Hutter, L.~Kotthoff, and J.~Vanschoren, Eds., \emph{Automated {{Machine
  Learning}}: {{Methods}}, {{Systems}}, {{Challenges}}}, ser. The {{Springer
  Series}} on {{Challenges}} in {{Machine Learning}}, 2019.

\bibitem{waringAutomatedMachineLearning2020}
J.~Waring, C.~Lindvall, and R.~Umeton, ``\BIBforeignlanguage{en}{Automated
  machine learning: {{Review}} of the state-of-the-art and opportunities for
  healthcare},'' \emph{\BIBforeignlanguage{en}{Artificial Intelligence in
  Medicine}}, vol. 104, p. 101822, Apr. 2020.

\bibitem{Garouani_2022}
M.~Garouani, M.~Hamlich, A.~Ahmad, M.~Bouneffa, G.~Bourguin, and
  A.~Lewandowski, ``Toward an~automatic assistance framework for~the~selection
  and~configuration of~machine learning based data analytics solutions
  in~industry 4.0,'' in \emph{Proceedings of the 5th International Conference
  on Big Data and Internet of Things}.\hskip 1em plus 0.5em minus 0.4em\relax
  Springer International Publishing, 2022, pp. 3--15.

\bibitem{Garouani2022-sf}
M.~Garouani, A.~Ahmad, M.~Bouneffa, M.~Hamlich, G.~Bourguin, and
  A.~Lewandowski, ``Towards meta-learning based data analytics to better assist
  the domain experts in industry 4.0,'' in \emph{Artificial Intelligence in
  Data and Big Data Processing}.\hskip 1em plus 0.5em minus 0.4em\relax Cham:
  Springer International Publishing, 2022, pp. 265--277.

\bibitem{vainshteinHybridApproachAutomatic2018a}
R.~Vainshtein, A.~Greenstein-Messica, G.~Katz, B.~Shapira, and L.~Rokach, ``A
  {{Hybrid Approach}} for {{Automatic Model Recommendation}},'' ser. {{CIKM}}
  '18, 2018, pp. 1623--1626.

\bibitem{10.1007/978-3-031-23615-0_33}
M.~Chaabi, M.~Hamlich, and \textbf{ M. Garouani}, ``Evaluation of automl tools
  for manufacturing applications,'' in \emph{Advances in Integrated Design and
  Production II}.\hskip 1em plus 0.5em minus 0.4em\relax Cham: Springer
  International Publishing, 2023, pp. 323--330.

\bibitem{Garouani2022-hm}
M.~Garouani and K.~Zaysa, ``Leveraging the automated machine learning for
  arabic opinion mining: A preliminary study on {AutoML} tools and comparison
  to human performance,'' in \emph{Digital Technologies and Applications}, ser.
  Lecture notes in networks and systems.\hskip 1em plus 0.5em minus 0.4em\relax
  Cham: Springer International Publishing, 2022, pp. 163--171.

\bibitem{kotthoffAutoWEKAAutomaticModel2019b}
L.~Kotthoff, C.~Thornton, H.~H. Hoos, F.~Hutter, and K.~Leyton-Brown,
  ``Auto-{{WEKA}}: {{Automatic Model Selection}} and {{Hyperparameter
  Optimization}} in {{WEKA}},'' in \emph{Automated {{Machine Learning}}:
  {{Methods}}, {{Systems}}, {{Challenges}}}, 2019, pp. 81--95.

\bibitem{olsonTPOTTreeBasedPipeline2019a}
R.~S. Olson and J.~H. Moore, ``\BIBforeignlanguage{en}{{{TPOT}}: {{A
  Tree}}-{{Based Pipeline Optimization Tool}} for {{Automating Machine
  Learning}}},'' in \emph{\BIBforeignlanguage{en}{Automated {{Machine
  Learning}}: {{Methods}}, {{Systems}}, {{Challenges}}}}, ser. The {{Springer
  Series}} on {{Challenges}} in {{Machine Learning}}, {Cham}, 2019, pp.
  151--160.

\bibitem{bilalliPRESISTANTDataPreprocessing2018a}
B.~Bilalli, A.~Abell\'o, T.~Aluja-Banet, R.~F. Munir, and R.~Wrembel,
  ``{{PRESISTANT}}: {{Data Pre}}-processing {{Assistant}},'' in
  \emph{Information {{Systems}} in the {{Big Data Era}}}, 2018, pp. 57--65.

\bibitem{feurerAutosklearnEfficientRobust2019}
M.~Feurer, A.~Klein, K.~Eggensperger, J.~T. Springenberg, M.~Blum, and
  F.~Hutter, ``Auto-sklearn: {{Efficient}} and {{Robust Automated Machine
  Learning}},'' in \emph{Automated {{Machine Learning}}: {{Methods}},
  {{Systems}}, {{Challenges}}}, ser. The {{Springer Series}} on {{Challenges}}
  in {{Machine Learning}}, F.~Hutter, L.~Kotthoff, and J.~Vanschoren, Eds.,
  2019, pp. 113--134.

\bibitem{kotuPredictiveAnalyticsData2014}
V.~Kotu and B.~Deshpande, \emph{Predictive Analytics and Data Mining: Concepts
  and Practice with RapidMiner}.\hskip 1em plus 0.5em minus 0.4em\relax {Morgan
  Kaufmann}, 2014.

\bibitem{H2OAiAI}
\BIBentryALTinterwordspacing
(2022, Dec.) {{H2O}}.ai | {{AI Cloud Platform}}. [Online]. Available:
  \url{https://www.h2o.ai/}
\BIBentrySTDinterwordspacing

\bibitem{BigML}
\BIBentryALTinterwordspacing
(2011) {{BigML}}. [Online]. Available: \url{https://bigml.com/}
\BIBentrySTDinterwordspacing

\bibitem{DataRobot}
\BIBentryALTinterwordspacing
(2012) {DataRobot | AI Cloud}. [Online]. Available:
  \url{https://www.datarobot.com/}
\BIBentrySTDinterwordspacing

\bibitem{automlchallenges}
I.~Guyon, L.~Sun-Hosoya, M.~Boull\'e, H.~J. Escalante, and al., ``Analysis of
  the {{AutoML Challenge Series}} 2015\textendash 2018,'' in \emph{Automated
  {{Machine Learning}}: {{Methods}}, {{Systems}}, {{Challenges}}}, ser. The
  {{Springer Series}} on {{Challenges}} in {{Machine Learning}}, 2019, pp.
  177--219.

\bibitem{bernsteinIntelligentAssistanceData2002}
A.~Bernstein, S.~Hill, and F.~Provost, ``Intelligent {{Assistance}} for the
  {{Data Mining Process}}: An {{Ontology-Based Approach}},'' 2002.

\bibitem{mustafavnuralOntologybasedSemanticsVs2017}
M.~V. Nural, ``Ontology-based semantics vs meta-learning for predictive big
  data analytics,'' 2017.

\bibitem{nuralUsingSemanticsPredictive2015}
M.~V. Nural, M.~E. Cotterell, and J.~A. Miller, ``Using {{Semantics}} in
  {{Predictive Big Data Analytics}},'' in \emph{2015 {{IEEE International
  Congress}} on {{Big Data}}}, 2015-06, pp. 254--261.

\bibitem{nuralUsingMetalearningModel2017a}
M.~V. Nural, H.~Peng, and J.~A. Miller, ``Using meta-learning for model type
  selection in predictive big data analytics,'' in \emph{2017 {{IEEE
  International Conference}} on {{Big Data}} ({{Big Data}})}, 2017, pp.
  2027--2036.

\bibitem{lakeBuildingMachinesThat2016}
B.~M. Lake, T.~D. Ullman, J.~B. Tenenbaum, and S.~J. Gershman, ``Building
  {{Machines That Learn}} and {{Think Like People}},'' 2016.

\bibitem{Garouani2022}
M.~Garouani, A.~Ahmad, M.~Bouneffa, M.~Hamlich, G.~Bourguin, and
  A.~Lewandowski, ``Using meta-learning for automated algorithms selection and
  configuration: an experimental framework for industrial big data,''
  \emph{Journal of Big Data}, vol.~9, no.~1, Apr. 2022.

\bibitem{vilaltaUsingMetaLearningSupport2004}
R.~Vilalta, C.~Giraud-Carrier, P.~Brazdil, and C.~Soares, ``Using
  {{Meta-Learning}} to {{Support Data Mining}},'' \emph{Int. J. Comput. Sci.
  Appl.}, 2004.

\bibitem{Brazdil2009}
P.~Brazdil, C.~Giraud-Carrier, C.~Soares, and R.~Vilalta, \emph{Metalearning:
  {{Applications}} to Data Mining}.\hskip 1em plus 0.5em minus 0.4em\relax
  {Springer Berlin Heidelberg}, 2009.

\bibitem{besimbilalliLearningImpactData2018}
B.~Bilalli, ``Learning the {{Impact}} of {{Data Pre-processing}} in {{Data
  Analysis}},'' 2018.

\bibitem{pfahringerMetaLearningLandmarkingVarious2001}
B.~Pfahringer, ``Meta-{{Learning}} by {{Landmarking Various Learning
  Algorithms}},'' 2001-05-23.

\bibitem{meskhiLearningAbstractTask2021}
M.~M. Meskhi, A.~Rivolli, R.~G. Mantovani, and R.~Vilalta, ``Learning
  {{Abstract Task Representations}},'' 2021-01-28.

\bibitem{garouani_iceis23}
M.~Garouani, A.~Ahmad, and M.~Bouneffa., ``Explaining meta-features importance
  in meta-learning through shapley values,'' in \emph{Proceedings of the 25th
  International Conference on Enterprise Information Systems - Volume 1:
  ICEIS,}, INSTICC.\hskip 1em plus 0.5em minus 0.4em\relax SciTePress, 2023,
  pp. 591--598.

\bibitem{rivolliCharacterizingClassificationDatasets2019}
A.~Rivolli, L.~P.~F. Garcia, C.~Soares, and J.~Vanschoren, ``Characterizing
  classification datasets: A study of meta-features for meta-learning,'' 2019.

\bibitem{reifAutomaticClassifierSelection2014}
M.~Reif, F.~Shafait, M.~Goldstein, T.~Breuel, and A.~Dengel, ``Automatic
  classifier selection for non-experts,'' \emph{Pattern Analysis and
  Applications}, vol.~17, no.~1, pp. 83--96, Feb. 2014.

\bibitem{castielloMetadataCharacterizationInput2005}
C.~Castiello, G.~Castellano, and A.~M. Fanelli, ``Meta-data:
  {{Characterization}} of {{Input Features}} for {{Meta-learning}},'' in
  \emph{Modeling {{Decisions}} for {{Artificial Intelligence}}}, ser. Lecture
  {{Notes}} in {{Computer Science}}.\hskip 1em plus 0.5em minus 0.4em\relax
  {Springer}, 2005, pp. 457--468.

\bibitem{kubaExploitingSamplingMetalearning2002}
P.~Kuba, P.~Brazdil, C.~Soares, and A.~Woznica, ``Exploiting {{Sampling}} and
  {{Meta-learning}} for {{Parameter Setting forSupport Vector Machines}},''
  2002.

\bibitem{kalousisModelSelectionMetalearning2000}
A.~Kalousis and M.~Hilario, ``Model selection via meta-learning: a comparative
  study,'' in \emph{Proceedings 12th {IEEE} Internationals Conference on Tools
  with Artificial Intelligence. {ICTAI} 2000}.\hskip 1em plus 0.5em minus
  0.4em\relax {IEEE} Comput. Soc, 2000.

\bibitem{gijsbersAutomaticConstructionMachine2017}
\BIBentryALTinterwordspacing
P.~Gijsbers, ``Automatic construction of machine learning pipelines,'' 2017-10.
  [Online]. Available:
  \url{https://research.tue.nl/en/studentTheses/automatic-construction-of-machine-learning-pipelines}
\BIBentrySTDinterwordspacing

\bibitem{furnkranzEvaluationLandmarkingVariants2001}
J.~F\"urnkranz and J.~Petrak, ``An {{Evaluation}} of {{Landmarking
  Variants}},'' 2001.

\bibitem{garciaNoiseDetectionMetalearning2016}
L.~P. Garcia, A.~C. de~Carvalho, and A.~C. Lorena, ``Noise detection in the
  meta-learning level,'' \emph{Neurocomputing}, vol. 176, pp. 14--25, feb 2016.

\bibitem{lorenaDataComplexityMetafeatures2018}
A.~C. Lorena, A.~I. Maciel, P.~B.~C. de~Miranda, I.~G. Costa, and R.~B.~C.
  Prud{\^{e}}ncio, ``Data complexity meta-features for regression problems,''
  \emph{Machine Learning}, vol. 107, no.~1, pp. 209--246, dec 2017.

\bibitem{hoComplexityMeasuresSupervised2002a}
T.~K. Ho and M.~Basu, ``Complexity measures of supervised classification
  problems,'' \emph{IEEE Transactions on Pattern Analysis and Machine
  Intelligence}, vol.~24, no.~3, pp. 289--300, 2002-03.

\bibitem{zhongguoChoosingClassificationAlgorithms2017}
Y.~Zhongguo, H.~Li, S.~Ali, and Y.~Ao, ``Choosing {{Classification Algorithms}}
  and {{Its Optimum Parameters}} based on {{Data Set Characteristics}},''
  \emph{Journal of Computers (Taiwan)}, vol.~28, pp. 26--38, 2017.

\bibitem{priyaUsingGeneticAlgorithms2012}
R.~Priya, B.~F. de~Souza, A.~L.~D. Rossi, and A.~de~Carvalho, ``Using genetic
  algorithms to improve prediction of execution times of {ML} tasks,'' in
  \emph{Lecture Notes in Computer Science}.\hskip 1em plus 0.5em minus
  0.4em\relax Springer Berlin Heidelberg, 2012, pp. 196--207.

\bibitem{pintoAutoBaggingLearningRank2017}
F.~Pinto, V.~Cerqueira, C.~Soares, and J.~a. Mendes-Moreira, ``{{autoBagging}}:
  {{Learning}} to {{Rank Bagging Workflows}} with {{Metalearning}},''
  2017-06-28.

\bibitem{molinaMetalearningApproachAutomatic2012}
M.~D.~M. Molina, C.~Romero, S.~Ventura, and J.~M. Luna, ``Meta-learning
  {{Approach}} for {{Automatic Parameter Tuning}}: {{A}} case of study with
  educational datasets,'' in \emph{{{Educational Data Mining}}}, 2012.

\bibitem{riddUsingMetalearningPredict2014}
P.~Ridd and C.~G. Giraud-Carrier, ``Using {{Metalearning}} to {{Predict When
  Parameter Optimization Is Likely}} to {{Improve Classification Accuracy}},''
  in \emph{Proceedings of the {{International Workshop}} on {{Meta-learning}}
  and {{Algorithm Selection}}}, ser. {{CEUR Workshop Proceedings}}, vol. 1201,
  2014, pp. 18--23.

\bibitem{Marini_2015}
\BIBentryALTinterwordspacing
F.~Marini and B.~Walczak, ``Particle swarm optimization ({PSO}). a tutorial,''
  \emph{Chemometrics and Intelligent Laboratory Systems}, vol. 149, pp.
  153--165, dec 2015. [Online]. Available:
  \url{https://doi.org/10.10162Fj.chemolab.2015.08.020}
\BIBentrySTDinterwordspacing

\bibitem{reifMetalearningEvolutionaryParameter2012}
M.~Reif, F.~Shafait, and A.~Dengel, ``Meta-learning for evolutionary parameter
  optimization of classifiers,'' \emph{Machine Learning}, vol.~87, no.~3, pp.
  357--380, 2012.

\bibitem{yangOBOECollaborativeFiltering2019}
C.~Yang, Y.~Akimoto, D.~W. Kim, and M.~Udell, ``{{OBOE}}: {{Collaborative
  Filtering}} for {{AutoML Model Selection}},'' \emph{Proceedings of the 25th
  ACM International Conference on Knowledge Discovery \& Data Mining}, pp.
  1173--1183, 2019.

\bibitem{bramerEstimatingPredictiveAccuracy2013}
M.~Bramer, ``Estimating the {{Predictive Accuracy}} of a {{Classifier}},'' in
  \emph{Principles of {{Data Mining}}}, ser. Undergraduate {{Topics}} in
  {{Computer Science}}, M.~Bramer, Ed.\hskip 1em plus 0.5em minus 0.4em\relax
  {Springer}, 2013, pp. 79--92.

\bibitem{guerraPredictingPerformanceLearning2008}
S.~B. Guerra, R.~B.~C. Prud\^encio, and T.~B. Ludermir, ``Predicting the
  {{Performance}} of {{Learning Algorithms Using Support Vector Machines}} as
  {{Meta-regressors}},'' in \emph{Artificial {{Neural Networks}} - {{ICANN}}
  2008}.\hskip 1em plus 0.5em minus 0.4em\relax {Springer}, 2008, pp. 523--532.

\bibitem{wistubaSequentialModelFreeHyperparameter2015}
M.~Wistuba, N.~Schilling, and L.~Schmidt-Thieme, ``Sequential model-free
  hyperparameter tuning,'' in \emph{2015 {IEEE} International Conference on
  Data Mining}.\hskip 1em plus 0.5em minus 0.4em\relax {IEEE}, nov 2015.

\bibitem{Garouani_2023}
M.~Garouani, A.~Ahmad, M.~Bouneffa, and M.~Hamlich, ``Autoencoder-{kNN}
  meta-model based data characterization approach for an automated selection of
  {AI} algorithms,'' \emph{Journal of Big Data}, vol.~10, no.~1, feb 2023.

\bibitem{wangFeatureSubsetSelection2013}
G.~Wang, Q.~Song, H.~Sun, X.~Zhang, B.~Xu, and Y.~Zhou, ``A feature subset
  selection algorithm automatic recommendation method,'' \emph{Journal of
  Artificial Intelligence Research}, vol.~47, pp. 1--34, may 2013.

\bibitem{reifPredictionClassifierTraining2011}
M.~Reif, F.~Shafait, and A.~Dengel, ``Prediction of {{Classifier Training Time
  Including Parameter Optimization}},'' in \emph{{{KI}} 2011: {{Advances}} in
  {{Artificial Intelligence}}}, ser. Lecture {{Notes}} in {{Computer Science}},
  J.~Bach and S.~Edelkamp, Eds.\hskip 1em plus 0.5em minus 0.4em\relax
  {Springer}, 2011, pp. 260--271.

\bibitem{munozInstanceSpacesMachine2018}
M.~A. Mu\~noz, L.~Villanova, D.~Baatar, and K.~Smith-Miles, ``Instance spaces
  for machine learning classification,'' \emph{Machine Learning}, vol. 107,
  no.~1, pp. 109--147, 2018-01-01.

\bibitem{garouaniUsingMetalearningAutomated2022}
M.~Garouani, A.~Ahmad, M.~Bouneffa, M.~Hamlich, G.~Bourguin, and
  A.~Lewandowski, ``Using meta-learning for automated algorithms selection and
  configuration: an experimental framework for industrial big data,''
  \emph{Journal of Big Data}, vol.~9, no.~1, apr 2022.

\bibitem{Ghareh_Mohammadi_2019}
F.~G. Mohammadi, H.~R. Arabnia, and M.~H. Amini, ``On parameter tuning in
  meta-learning for computer vision,'' in \emph{2019 International Conference
  on Computational Science and Computational Intelligence ({CSCI})}.\hskip 1em
  plus 0.5em minus 0.4em\relax {IEEE}, dec 2019.

\bibitem{Arredondo_2015}
T.~Arredondo and W.~Ormaz{\'{a}}bal, ``Meta-learning framework applied in
  bioinformatics inference system design,'' \emph{International Journal of Data
  Mining and Bioinformatics}, vol.~11, no.~2, p. 139, 2015.

\bibitem{Smith_Miles_2009}
K.~A. Smith-Miles, ``Cross-disciplinary perspectives on meta-learning for
  algorithm selection,'' \emph{{ACM} Computing Surveys}, vol.~41, no.~1, pp.
  1--25, jan 2009.

\bibitem{2004.11149}
\BIBentryALTinterwordspacing
H.~Peng, ``A comprehensive overview and survey of recent advances in
  meta-learning,'' 2020. [Online]. Available:
  \url{https://arxiv.org/abs/2004.11149}
\BIBentrySTDinterwordspacing

\bibitem{thorntonAutoWEKACombinedSelection2013}
C.~Thornton, F.~Hutter, H.~H. Hoos, and K.~Leyton-Brown, ``Auto-{{WEKA}}:
  Combined selection and hyperparameter optimization of classification
  algorithms,'' in \emph{Proceedings of the 19th {{ACM SIGKDD}} International
  Conference on {{Knowledge}} Discovery and Data Mining}, ser. {{KDD}} '13,
  2013-08-11, pp. 847--855.

\bibitem{AutoMLTables}
\BIBentryALTinterwordspacing
(2011) Google cloud. [Online]. Available: \url{https://cloud.google.com/automl}
\BIBentrySTDinterwordspacing

\bibitem{jinAutoKerasEfficientNeural2019a}
\BIBentryALTinterwordspacing
H.~Jin, Q.~Song, and X.~Hu, ``Auto-{{Keras}}: {{An Efficient Neural
  Architecture Search System}},'' 2019. [Online]. Available:
  \url{http://arxiv.org/abs/1806.10282}
\BIBentrySTDinterwordspacing

\bibitem{8257923}
T.~Swearingen, W.~Drevo, B.~Cyphers, A.~Cuesta-Infante, A.~Ross, and
  K.~Veeramachaneni, ``Atm: A distributed, collaborative, scalable system for
  automated machine learning,'' in \emph{2017 IEEE International Conference on
  Big Data (Big Data)}, 2017, pp. 151--162.

\bibitem{ianh.DataMiningPractical2016}
W.~Ian~H., F.~Eibe, and H.~Mark~A., \emph{Data {{Mining}}: {{Practical Machine
  Learning Tools}} and {{Techniques}}}, 4th~ed.\hskip 1em plus 0.5em minus
  0.4em\relax {Morgan Kaufmann}, 2016.

\bibitem{vanschorenOpenMLNetworkedScience2014}
J.~Vanschoren, J.~N. van Rijn, B.~Bischl, and L.~Torgo, ``{OpenML},''
  \emph{{ACM} {SIGKDD} Explorations Newsletter}, vol.~15, no.~2, pp. 49--60,
  jun 2014.

\bibitem{alcobacaMFEReproducibleMetafeature2020}
E.~Alcobaça, F.~Siqueira, and A.~Rivolli, ``Mfe: Towards reproducible
  meta-feature extraction,'' \emph{Journal of Machine Learning Research},
  vol.~21, no. 111, pp. 1--5, 2020.

\bibitem{gijsbersOpenSourceAutoML2019}
P.~Gijsbers, E.~LeDell, J.~Thomas, S.~Poirier, B.~Bischl, and J.~Vanschoren,
  ``An open source automl benchmark,'' 2019.

\bibitem{mazumderFailureRiskAnalysis2021b}
R.~K. Mazumder, A.~M. Salman, and Y.~Li, ``Failure risk analysis of pipelines
  using data-driven machine learning algorithms,'' \emph{Structural Safety},
  vol.~89, p. 102047, 2021.

\bibitem{saravanamuruganChatterPredictionBoring2017}
S.~Saravanamurugan, S.~Thiyagu, N.~R. Sakthivel, and B.~Nair,
  ``\BIBforeignlanguage{en}{Chatter prediction in boring process using machine
  learning technique},'' \emph{\BIBforeignlanguage{en}{Int. J. Manuf. Res.}},
  2017.

\bibitem{benkedjouhHealthAssessmentLife2015}
T.~Benkedjouh, K.~Medjaher, N.~Zerhouni, and S.~Rechak,
  ``\BIBforeignlanguage{en}{Health assessment and life prediction of cutting
  tools based on support vector regression},''
  \emph{\BIBforeignlanguage{en}{Journal of Intelligent Manufacturing}},
  vol.~26, no.~2, pp. 213--223, Apr. 2015.

\bibitem{garouaniAutomationIndustrialData2021b}
M.~Garouani, A.~Ahmad, M.~Bouneffa, A.~Lewandowski, G.~Bourguin, and
  M.~Hamlich, ``Towards the {{Automation}} of {{Industrial Data Science}}: {{A
  Meta}}-learning based {{Approach}},'' 2021, pp. 709--716.

\bibitem{vanschorenMetaLearningSurvey2018}
J.~Vanschoren, ``Meta-{{Learning}}: {{A Survey}},'' 2018-10-08.

\bibitem{matejkaSameStatsDifferent2017}
J.~Matejka and G.~Fitzmaurice, ``Same {{Stats}}, {{Different Graphs}}:
  {{Generating Datasets}} with {{Varied Appearance}} and {{Identical
  Statistics}} through {{Simulated Annealing}},'' in \emph{Proceedings of the
  2017 {{CHI Conference}} on {{Human Factors}} in {{Computing Systems}}}, 2017,
  pp. 1290--1294.

\bibitem{amlbid}
M.~Garouani, A.~Ahmad, M.~Bouneffa, and M.~Hamlich, ``{AMLBID}: An
  auto-explained automated machine learning tool for big industrial data,''
  \emph{{SoftwareX}}, vol.~17, p. 100919, Jan. 2022.

\end{thebibliography}

\newpage

% \section{Biography}

% \begin{IEEEbiography}[{\includegraphics[width=1in,height=2.25in,clip,keepaspectratio]{Figures/moncef}}]{Moncef Garouani}
%  is an Associate Professor at the University of Toulouse Capitole and a member of the IRIT laboratory. He received his PhD in Computer Science from the University of Littoral Côte d’Opale and Hassan II University in 2022. His research interests encompass artificial intelligence, big data analysis, meta-learning, and explainable artificial intelligence.
% \end{IEEEbiography}

% \begin{IEEEbiography}[{\includegraphics[width=1in,height=2.25in,clip,keepaspectratio]{Figures/ayah}}]{Ayah Barhrhouj }
% is currently pursuing a PhD degree in the LIS laboratory at Aix-Marseille University in France. Her research interests include machine learning, artificial intelligence, and explainable artificial intelligence.
% \end{IEEEbiography}

% \begin{IEEEbiography}[{\includegraphics[width=1in,height=2.25in,clip,keepaspectratio]{Figures/mourad}}]{Mourad Bouneffa }
% is a Full Professor at the University of Littoral Côte d’Opale. His research interests include big data analysis, meta-learning, and explainable artificial intelligence.
% \end{IEEEbiography}

% \begin{IEEEbiography}[{\includegraphics[width=1in,height=2.25in,clip,keepaspectratio]{Figures/adeel}}]{Adeel Ahmad }
% is an Associate Professor at the University of Littoral Côte d’Opale. His research interests include big data analysis.

% \end{IEEEbiography}

\vfill

\end{document}